\documentclass[sigconf,authorcopy,screen]{acmart}

\renewcommand\footnotetextcopyrightpermission[1]{} 
\usepackage{graphicx, caption, subcaption}
\graphicspath{ {./images/}}
\AtBeginDocument{%
  \providecommand\BibTeX{{%
    \normalfont B\kern-0.5em{\scshape i\kern-0.25em b}\kern-0.8em\TeX}}}
\setcopyright{none}

\begin{document}
\title{Assessing Robustness of Deep learning Methods\\ 
in Dermatological Workflow}

\author{Sourav Mishra}
\email{sourav@ay-lab.org}
\affiliation{%
  \institution{The University of Tokyo}
  \streetaddress{7-3-1 Hongo, Bunkyo-ku}
  \city{Tokyo}
}
\author{Subhajit Chaudhury}
\email{subhajit@ay-lab.org}
\affiliation{%
  \institution{The University of Tokyo}
  \streetaddress{7-3-1 Hongo, Bunkyo-ku}
  \city{Tokyo}
}
\author{Hideaki Imaizumi}
\email{imaq@exmed.io}
\affiliation{%
  \institution{ExMedio Inc.}
  \streetaddress{3-5-1 Kojimachi, Chiyoda-ku}
  \city{Tokyo}
}
\author{Toshihiko Yamasaki}
\email{yamasaki@ay-lab.org}
\affiliation{%
  \institution{The University of Tokyo}
  \streetaddress{7-3-1 Hongo, Bunkyo-ku}
  \city{Tokyo}
}
\renewcommand{\shortauthors}{Mishra et. al}

\begin{abstract}
\label{abstract}
This paper aims to evaluate the suitability of current deep 
learning methods for clinical workflow especially by focusing on 
dermatology. Although deep learning methods have been attempted to 
get dermatologist level accuracy in several individual conditions, 
it has not been rigorously tested for common clinical complaints. 
Most projects involve data acquired in well-controlled laboratory 
conditions. This may not reflect regular clinical evaluation where 
corresponding image quality is not always ideal. We test the robustness 
of deep learning methods by simulating non-ideal characteristics on user 
submitted images of ten classes of diseases. Assessing via imitated 
conditions, we have found the overall accuracy to drop and individual 
predictions change significantly in many cases despite of robust 
training.
\end{abstract}

\ccsdesc[500]{Computing methodologies~Learning paradigms}
\ccsdesc[500]{Applied computing~Health care information systems}

\keywords{Deep learning, neural networks, 
dermatology, robustness, inference}

\maketitle

\section{Introduction}
\label{introduction}
Deep learning (DL) has found increasing use in healthcare domain 
in recent years. With its ability to find patterns imperceptible 
to human evaluation, it promises to be a powerful tool in improving 
the quality of treatment. The Food and Drug Agency (FDA) of the US has 
recently approved diagnostic tools such as \textit{QuantX} and 
\textit{EchoMD AutoEF}, which significantly improve diagnosing 
breast cancer and cardiac abnormalities 
respectively~\cite{allen2019role}. DeepMind is collaborating 
with National Health Scheme (NHS) of UK to streamline 
patient care~\cite{powles2017google}. 
Quality dermatological attention is also an established need in 
the current scenario. Approximately 1.9 billion people worldwide 
are having some skin conditions, with skin diseases being the 
fourth most common source of morbidity~\cite{liu2019deep}. 
But there is a shortage of dermatologists to treat this 
growing patient pool. In the US, there are only 3.6 
dermatologists reported for every 10,000 people~\cite{kimball2008us}. 
Canada and Japan are courting telemdicine as a remedial means 
to provide specialist opinion in remote 
areas~\cite{lanzini2012impact,dekio2010,imaizumi2017}.

The variety of skin conditions is fairly large. Diseases such as 
contact dermatitis and ringworm, though not life threatening, 
spread virulently. According to the estimates by National Institutes 
of Health (NIH) in the US, one out of five US citizens are at a risk 
for morbidity due to debilitating skin conditions~\cite{stern2010}. 
With timely intervention many of these conditions can be resolved. 
The survival rate is close to 98\% with remedial therapeutics.

At a time when the need for dermatological expertise is increasing 
due to growing incidence, a constant undersupply of specialists 
is leading to long waiting queues~\cite{mishra2018a, suneja2001}. 
In the absence of immediate specialist attention, patients tend 
to consult general practitioners. Their diagnostic accuracy is 
reported to be between 24-70\% and concurrency with specialist 
opinion around 57\%~\cite{liu2019deep,lowell2001}. Specialist 
opinion can also vary from person to person, leading to increased 
risk to the patient. 

Computer vision has gained traction in several domains due to the 
advent of deep learning~\cite{lecun2015DL}. They have circumvented 
rule based approaches, which used to make identification of visual 
patterns tedious. There have been few successful case studies in 
dermatology using deep learning. Esteva et al. and Haenssle et al. 
demonstrated that deep learning could match or even surpass 
dermatologists in detecting 
\textit{Melanoma}~\cite{esteva2017,haenssle2018}. 
Shrivastava et al. could demonstrate similar efficacy of DL based 
\textit{Psoriasis} detection~\cite{shrivastava2015}. These models 
were limited to identifying single diseases. Park et al. could 
detect several anomalies with the help of crowd-sourcing~\cite{park2018}. 
Liu et al. have managed to achieve about 70\% accuracy over 26 
skin conditions by differential diagnosis~\cite{liu2019deep}. 

At present, we require some automation via expert systems to 
address the growing requirement of specialist attention. Although 
there have been recent forays into building such systems, the 
results are based on carefully curated datasets 
~\cite{liu2019deep,mishra2019interpreting,codella2017}. In trying 
to perform machine aided diagnostics, we believe handpicked 
multimedia obtained under ideal conditions are not a true 
representation of samples seen in clinical workflow. In this 
paper, we attempt to assess the robustness of deep learning 
models, which have been well trained on user-submitted images 
over ten classes. We evaluate its performance on common workflow 
issues such as noise and blur. The paper is organized as follows. 
Section~2 introduces the data and methods which have been employed 
in the current study. Section~3 elaborates on the results. We 
present a brief discussion and summary in Section~4 before 
concluding. The contribution of this paper is as follows:
\begin{itemize}
    \item We have attempted to understand a multi-class dermatological 
    classifier's decision on wrong predictions via attention based 
    mechanisms.
    \item We have investigated the effect of common artifacts such 
    as shot noise and motion blur in changing the decisions.
    \item We have assessed the performance of such classifiers in 
    the presence of distribution shift and out-of-distribution 
    samples. 
    \item The source codes will be made publicly available so 
    that researchers in medical image recognition can evaluate the 
    robustness of their methods to such non-ideal conditions.
\end{itemize}

\section{Methods}
\subsection{Data}
\label{data}
A systematic collection of dermatological images was made with 
the consent and cooperation of volunteers belonging to the East 
Asian race. These images came from different sources, but usually 
larger than $200\times200$ pixels and in JPEG format. Some 
additional samples were sourced from medical centers and 
affiliated institutions within agreed frameworks of data reuse. 
These samples were anonymized and labeled by registered clinicians 
without any modifications. Photographs with identifying features 
such as face, birthmark, tattoos, hospital tags etc. were excluded 
from the study. With the advice of clinical practitioners, we chose 
the following ten classes to build and investigate our classifier:
(i)~Acne, (ii)~Alopecia, (iii)~Blister, (iv)~Crust, (v)~Erythema, 
(vi)~Leukoderma, (vii)~Pigmented Maculae, (viii)~Tumor, (ix)~Ulcer, 
and (x)~Wheal. Table~\ref{tab:exmd-data} presents information about 
the selected labels and their sizes.

\begin{table}[t]
\caption{Distribution of samples over selected labels.}
\label{tab:exmd-data}
\begin{tabular}{lc}
\toprule
    Label/Class & Samples\\
\midrule
    Acne        & 971\\
    Alopecia    & 681\\
    Blister     & 690\\
    Crust       & 639\\
    Erythema    & 689\\
    Leukoderma  & 664\\
    P. Macula   & 717\\
    Tumor       & 790\\
    Ulcer       & 782\\
    Wheal       & 636\\
  \bottomrule
\end{tabular}
\end{table}

To examine the effect of distribution shift and out of distribution 
samples, we chose the SD-198 dataset~\cite{sun2016}. It was an ideal 
candidate in our study since it was also composed of user submitted 
images. Since a one-to-one correspondence did not exist between 
classes in this dataset and our collected images, a dermatologist 
helped group classes relevant to our experimental design. Hundred 
samples were selected at random from these composite new classes 
and tested with our model. Information about this grouping is 
shown in Table~\ref{tab:sd198}.

\begin{table}[t]
\caption{Grouping of SD-198 classes relevant to study.}
\label{tab:sd198}
\begin{tabular}{ll}
\toprule
    Label & SD-198 Classes\\
\midrule
Acne        &   Acne Keloidalis Nuchae, Acne Vulgaris, Steroid  \\ 
            &   Acne, Favre Racouchot, Nevus Comedonicus,       \\
            &   Pomade Acne                                     \\
Alopecia    &   Alopecia Areata, Androgenetic Alopecia,         \\
            &   Follicular Mucinosis, Kerion, Scar Alopecia.    \\
Blister     &   Dyshidrosiform Eczema, Hailey Disease,          \\
            &   Herpes Simplex, Herpes Zoster, Varicella,       \\
            &   Mucha Habermann disease                         \\
Crust       &   Angular Cheilitis, Bowen's Disease, Impetigo    \\
Erythema    &   Acute Eczema, Candidiasis, Erythema Ab Igne,    \\
            &   Ery. Annulare Centrifigum, Ery. Craqule, Ery.   \\
            &   Multiforme, Rosacea, Exfoliative Erythroderma   \\
Leukoderma  &   Balanitis Xerotica Obl., Beau's Lines, Halo     \\
            &   Nevus, Leukonychia, Pityriasis Alba, Vitiligo   \\
P. Macula   &   Actinic Solar Damage, Becker's Nevus, Blue      \\   
            &   Nevus, Cafe Au Lait Macula, Compound Nevus,     \\
            &   Congenital Nevus Dermatosis Nigra, Epidermal    \\
            &   Nevus, Green Nail                               \\
Tumor       &   Angioma, Apocrine Hydrocystoma, Lipoma,         \\
            &   Dermatofibroma, Digital Fibroma, Fibroma,       \\
            &   Leiomyoma                                       \\
Ulcer       &   Aphthous Ulcer, Behcet's Disease, Ulcer, Stasis \\
            &   Ulcer, Mal Perforans, Pyoderma Gangrenosum,     \\
            &   Syringoma                                       \\
Wheal       &   Urticaria, Stasis Edema                         \\
\bottomrule
\end{tabular}
\end{table}

\subsection{Model learning}
\label{model-learning}
Since the dataset is much smaller as compared to conventional computer 
vision datasets, we chose previously established learning paradigms for 
rapid training as our starting point~\cite{mishra2019interpreting,
mishra2019improving}. We provide a brief overview of the process 
in this section. The model was built on PyTorch using a single GPU 
(NVIDIA\textsuperscript{TM} V100 16GB HBM2)~\cite{pytorch2019}. 
Pre-trained ResNet-34, ResNet-50, ResNet-101, and ResNet-152 were 
chosen as the candidate architectures since they exhibit feature 
re-use and propagation, which were essential to our fine tuning. We 
normalized data with the recommended mean and standard deviation. 
The data was split in 5:1 ratio into training and validation sets. 
We performed dynamic in-memory augmentation such as crop, random zoom, 
horizontal \& vertical flips in the data-loader.

\begin{figure}
\begin{minipage}[t]{\linewidth}
    \begin{center}
        \includegraphics[width=0.99\linewidth,trim={0cm 0cm 1cm 1cm}, 
    clip]{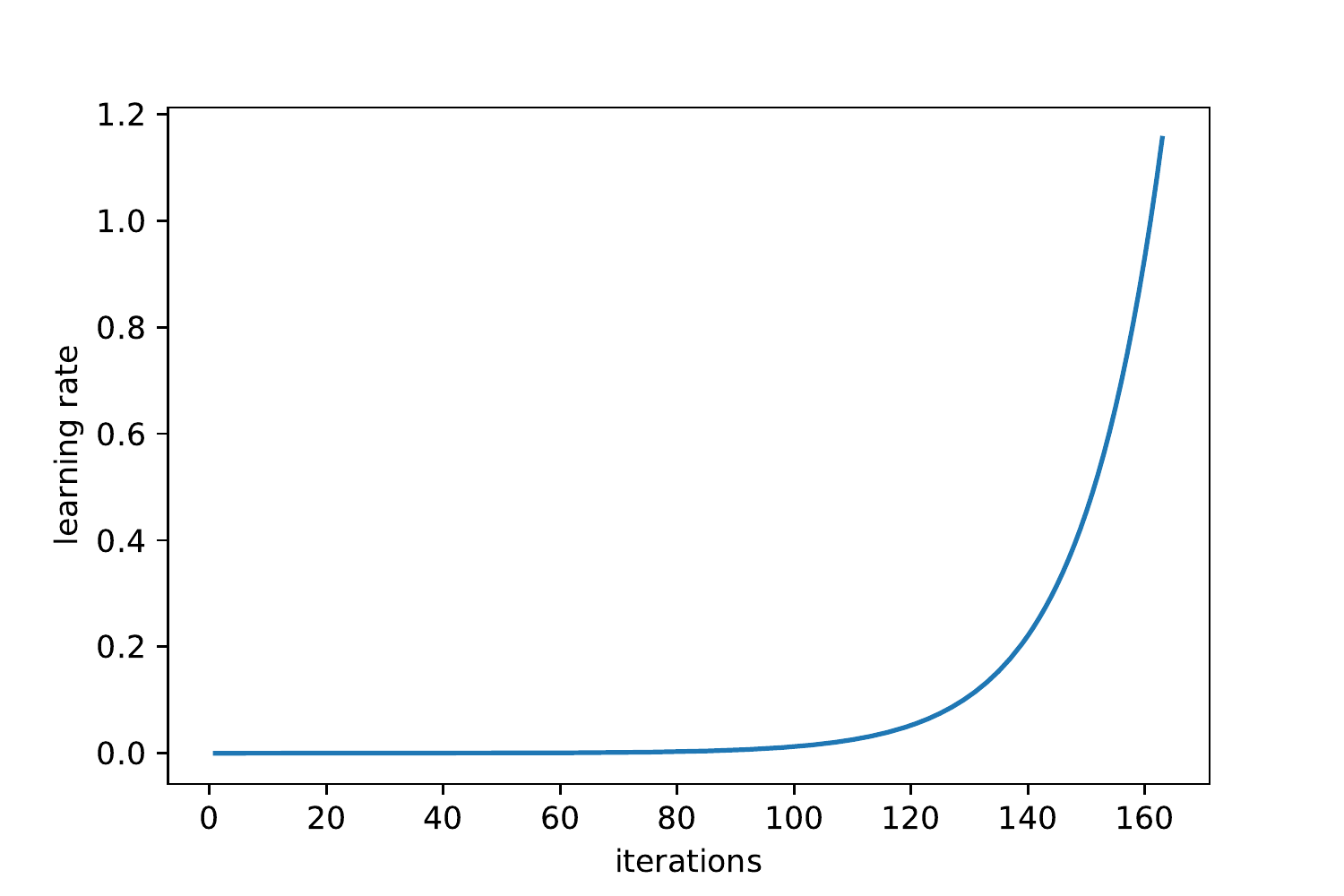}
    \end{center}
    \caption{LR ($\alpha$) was gradually increased over several 
    min-batches to observe the validation losses.}
    \label{fig:LR1}

  \begin{center}
    \includegraphics[width=0.99\linewidth,trim={0cm 0cm 1cm 1cm}, 
    clip]{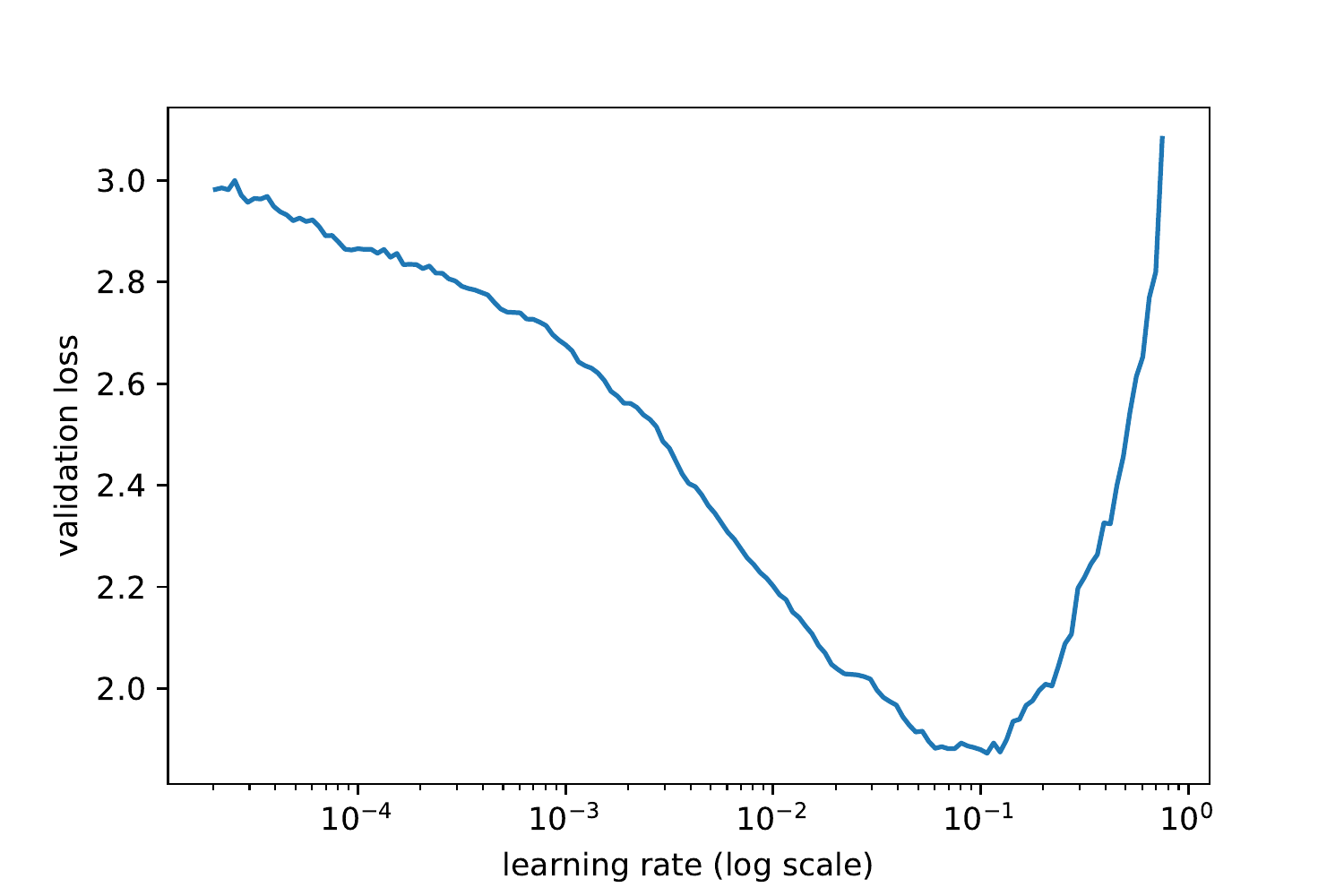}
  \end{center}
    \caption{Losses decreased typically with increasing 
    $\alpha$ until an inflexion point; The optimal value belonged in this 
    region.}
    \label{fig:LR2}
\end{minipage}
\end{figure}

For a good learning rate (LR), we calculated suitable values
through a range test prior to the model training~\cite{smith2017}. 
The implementation used several mini-batches with increasing values 
of the rate $\alpha$, and the loss values were computed. The validation
loss was observed until it dropped significantly and reached a point of 
inflexion. The learning rate was chosen in the neighborhood of this 
inflexion. Figures~\ref{fig:LR1} and \ref{fig:LR2} illustrate this 
process.

\begin{figure}
\begin{minipage}[t]{.45\textwidth}
    \begin{center}
    \includegraphics[width=0.99\linewidth,trim={0cm 0cm 1cm 1cm}, 
    clip]{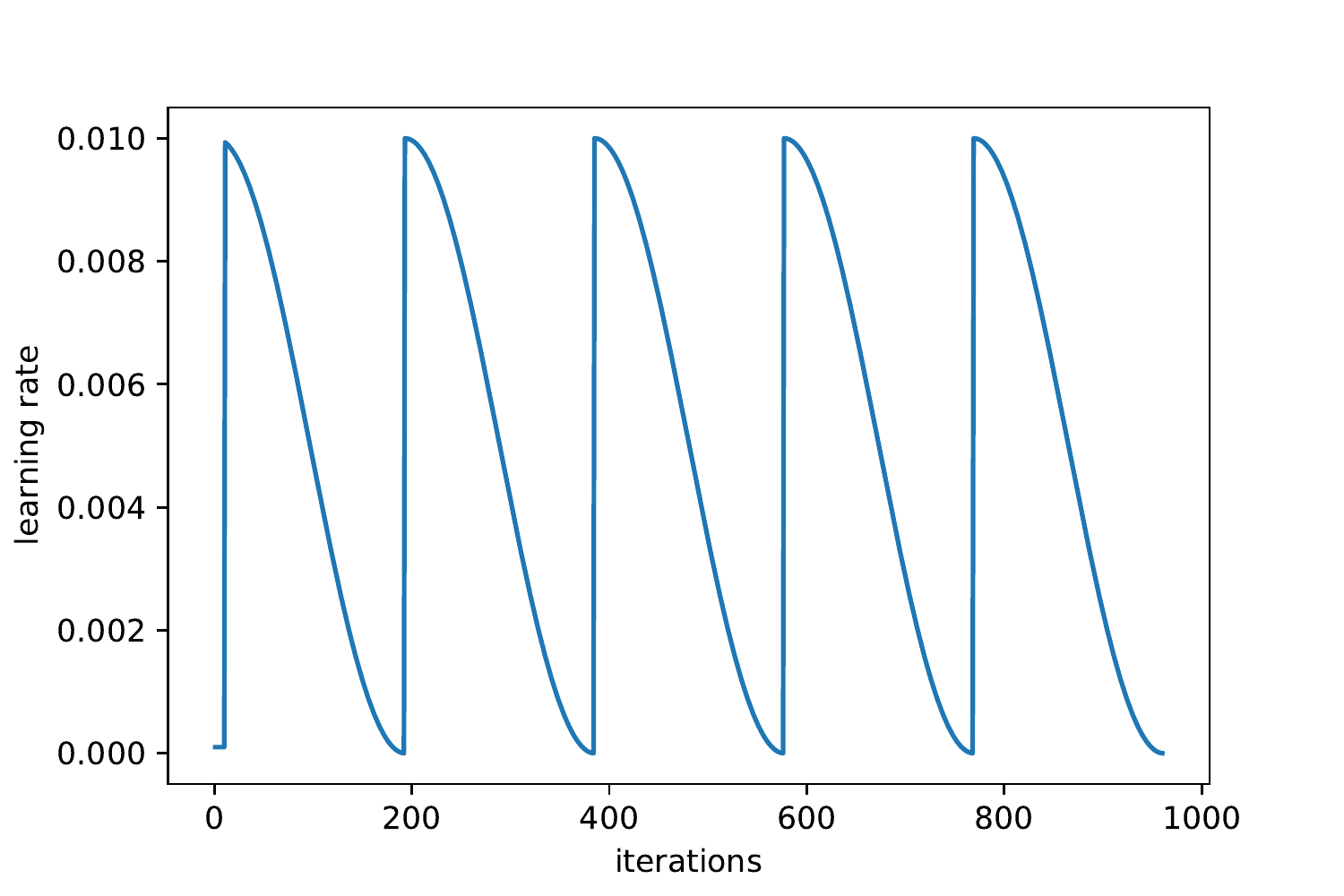}
    \end{center}
    \caption{Cyclical LR was used to train the final FC layer in 
    first phase of the model learning which resulted in coarse fits.}
    \label{fig:SGDR1}

    \begin{center}
    \includegraphics[width=0.99\linewidth,trim={0cm 0cm 1cm 1cm}, 
    clip]{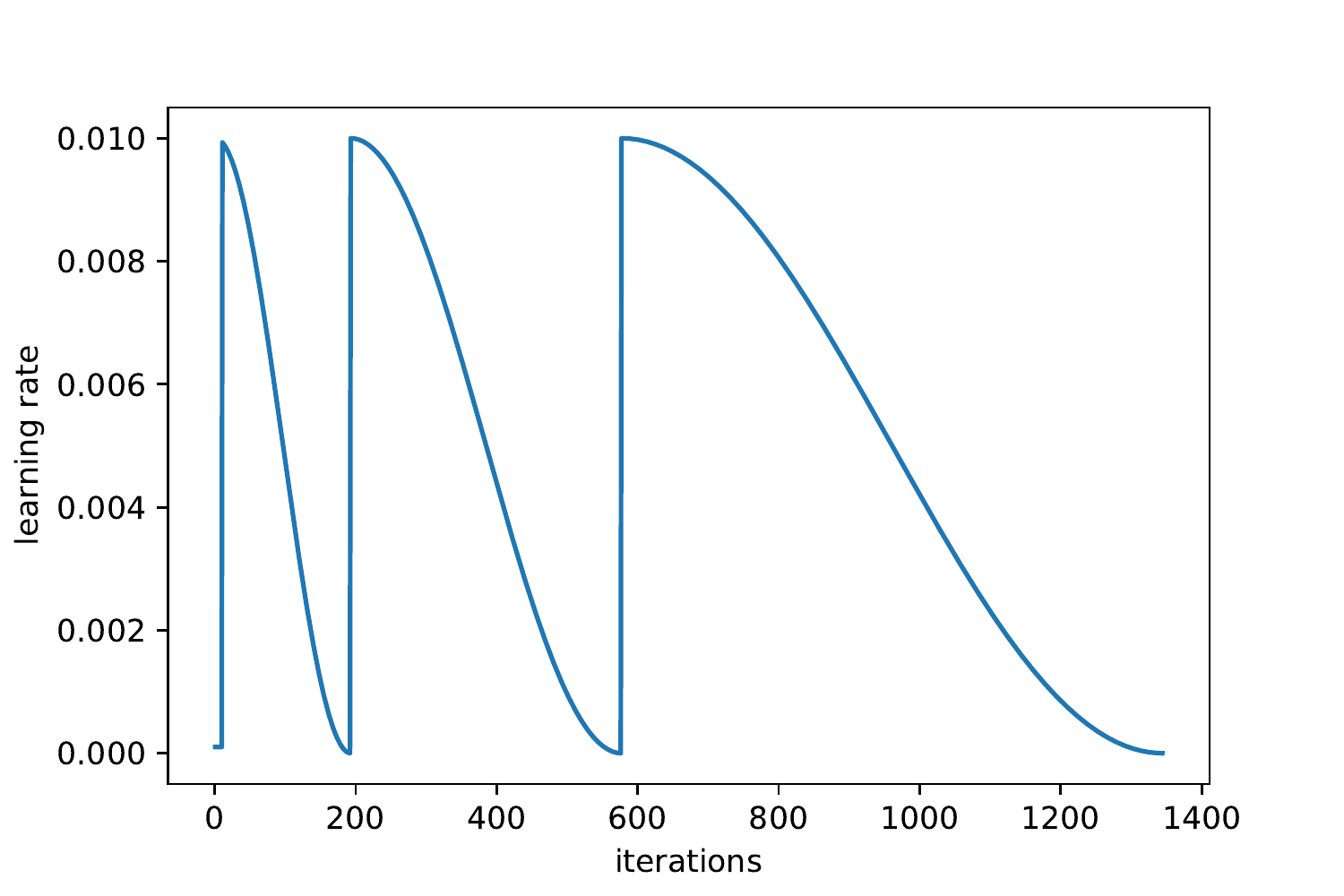}
    \end{center}
    \caption{Cycle length multiplication extended the rate cycling 
    over increasing number of epochs to prevent undesirable model 
    perturbation.}
    \label{fig:SGDR2}
    \end{minipage}
\end{figure}

The model learning was conducted in two phases using Stochastic 
gradient descent with warm restarts (SGD-R)~\cite{loshchilov2016}.
The first phase used a cyclical learning rate with the network 
frozen except the final fully-connected layer. Cosine rate annealing 
was performed at the beginning of each epoch. The LR modulation was 
governed by,
\begin{equation}
        \nu_t = \frac{1}{2} \left(1 + \nu 
        cos\left(\frac{t\pi}{T}\right)\right) 
        + \epsilon,
\label{eq:sgdr}
\end{equation}
where $\nu$ indicates the initial LR, $t$ is the current iteration, 
and $T$ is the total number of iterations to cover an epoch. 
This scheduling operation is illustrated in Figure~\ref{fig:SGDR1}.

In the subsequent phase, all the layers were unfrozen and SGD 
with restarts was used to train the full network. Two additional 
changes were also introduced: SGD-R was used in conjunction with 
cycle length multiplication (CLM) such that each cosine annealing 
cycle lasted longer. The modified SGD-R scheme is illustated in 
Figure~\ref{fig:SGDR2}. Additionally, discrimative learning 
rate was used to assign different scales of changes for different 
parts of the network~\cite{howard2018universal}. Doing so helped 
preserve valuable existing pre-trained information. The outcomes 
of our model learning are discussed in Sec.~\ref{result-learning}.

\subsection{Adversarial tests}
\label{adversarial-test}
Imaging systems are prone to imperfections. Shot noise can manifest 
as a result of CCD sensor's defects or via spots on the camera lens. 
Presence of noisy pixels can adversely affect the classifier 
performance~\cite{goodfellow2014explaining}. Unless using a stable 
platform for image capture, a photograph can also present varying 
amounts of motion blur. Sometimes both the artifacts could manifest 
in the image. 

When working with user submitted images, it is not unreasonable 
to expect these imperfections. Dermatological classifiers in clinical 
workflow should ideally be able to disregard the presence of 
minor flaws towards reliably identifying the skin lesions. 
Although benchmarks for adversarial robustness do not exist for the 
current application, we attempted to simulate the above mentioned 
artifacts on our dataset and compare the changes in accuracy.

\begin{figure}
  \begin{minipage}{\linewidth}
    \centering
    \subcaptionbox{Original}
    {\includegraphics[width=0.45\linewidth]{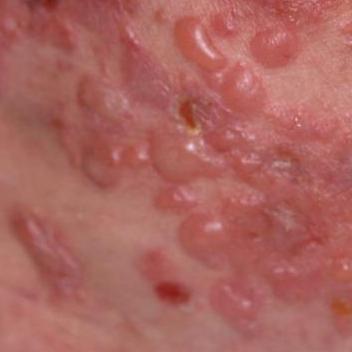}}\quad
    \subcaptionbox{S\&P addition}
    {\includegraphics[width=0.45\linewidth]{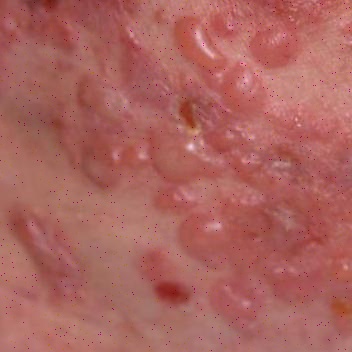}}
    \caption[Example of S\&P corruption]{Example of S\&P corruption}
    \label{fig:sp_corruption}
  \end{minipage}
\end{figure}

\begin{figure}
  \begin{minipage}{\linewidth}
    \centering
    \subcaptionbox{Original}
    {\includegraphics[width=0.45\linewidth]{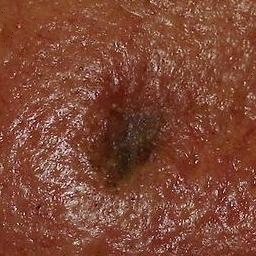}}\quad
    \subcaptionbox{Gaussian blurred}
    {\includegraphics[width=0.45\linewidth]{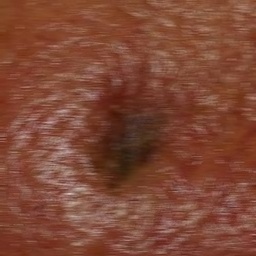}}
    \caption[Motion blur imitation]{Example of motion blur imitation}
    \label{fig:blur_corruption}
  \end{minipage}
\end{figure}

To simulate the effect of shot noise, we introduced varying levels 
of salt and pepper (S\&P) pixels. At most, 5\% of the image pixels 
were to be changed with the ratio of black to white noise pixels set 
at 3:1. In the case of motion blur, we chose a $5\times5$ Gaussian 
kernel. Sample images demonstrating the degree of changes are shown 
in Figures~\ref{fig:sp_corruption} and \ref{fig:blur_corruption}.

We tested the effects of these artifacts independent of each other. 
Additionally, we trained our classifier on noisy data to assess any 
change in the performance. The results of these adversarial 
experiments are discussed in Sec.~\ref{result-artifacts}.

\section{Results}
\label{results}
\subsection{Model fitting }
\label{result-learning}
\begin{figure}[t]
    \includegraphics[width=\linewidth, trim={1cm 0cm 1cm 1cm}, 
    clip]{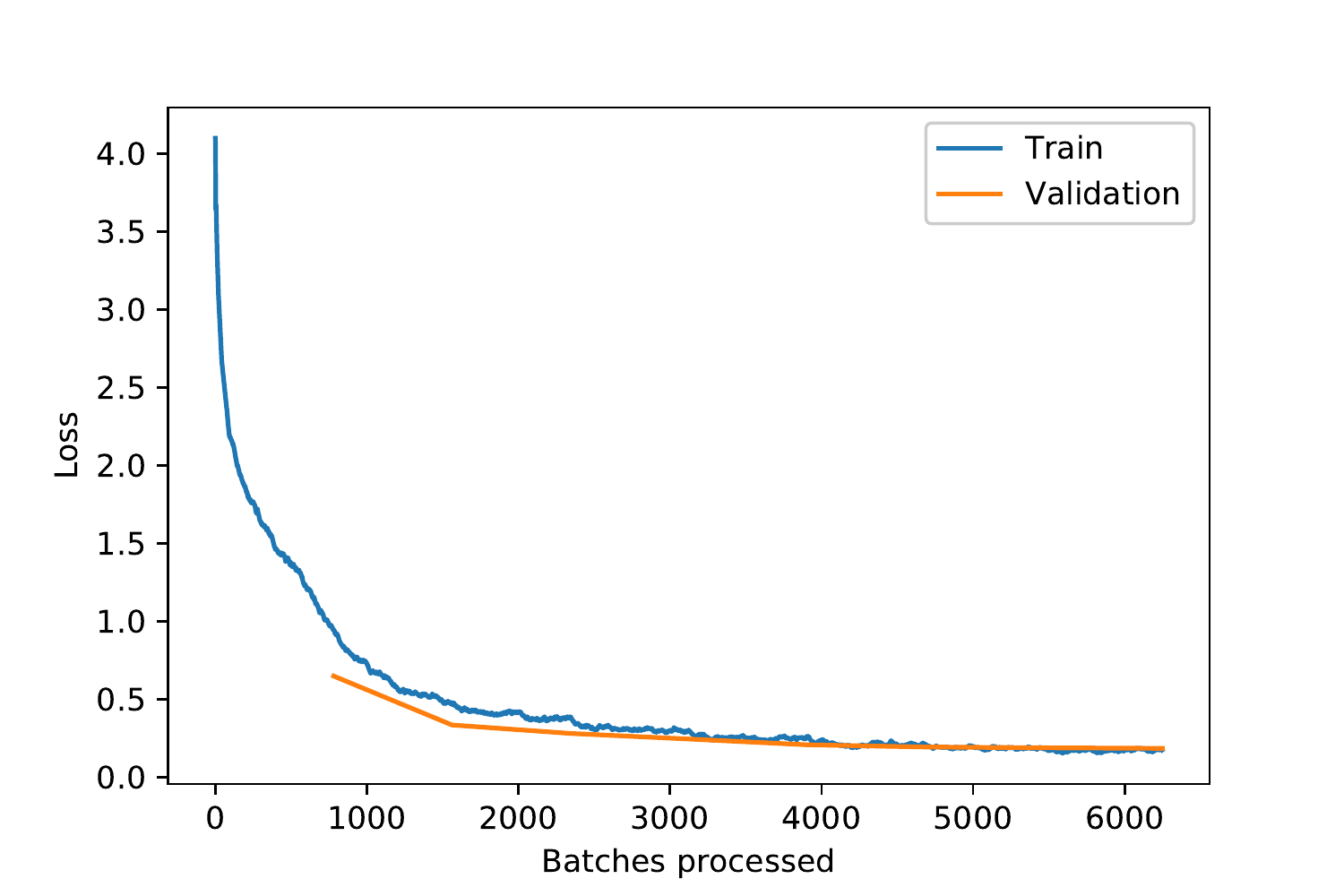}
    \caption{Error curves for model learning on ResNet-152}
    \label{fig:learning-curves}
\end{figure}

\begin{figure}[t]
    \includegraphics[width=\linewidth]{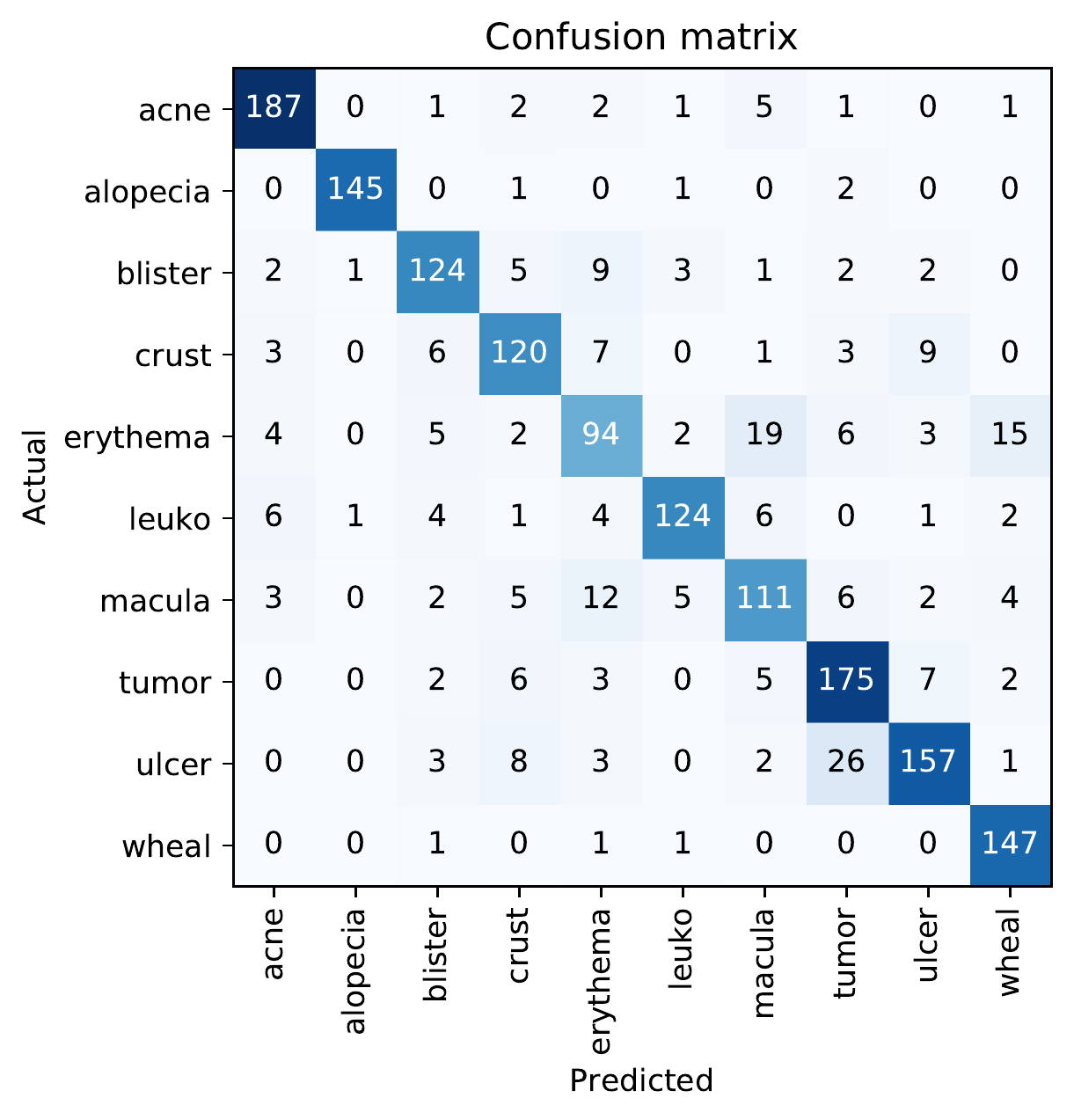}
    \caption{Confusion matrix for ResNet-152 model learning}
    \label{fig:exmd-confusion-R152}
\end{figure}

\begin{figure*}
  \begin{minipage}[!htb][][t]{\linewidth}
    \centering
    \subcaptionbox{Acne}
    {\includegraphics[width=0.30\linewidth, trim={0.5cm 0cm 1.5cm 0.5cm}, 
    clip]{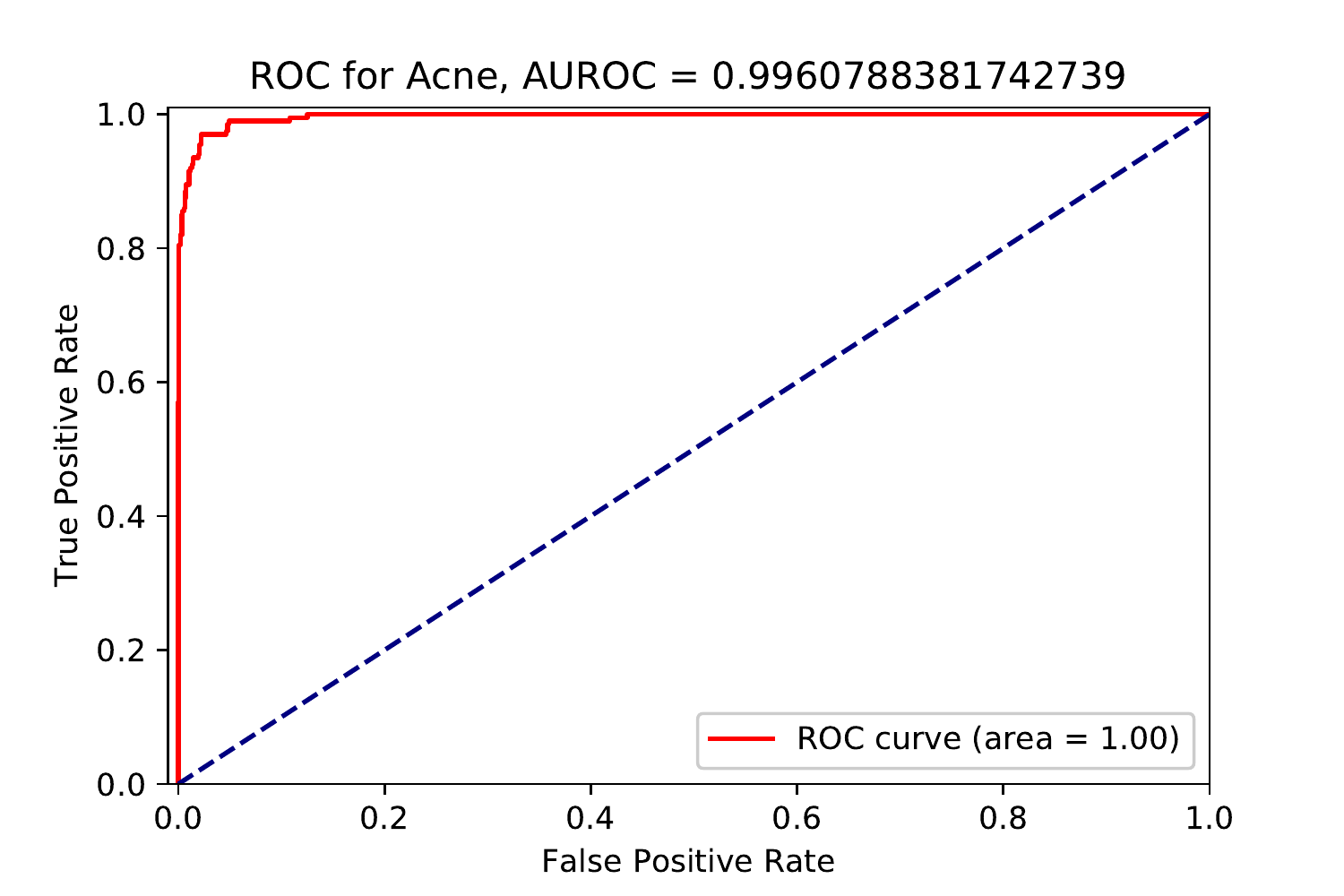}}\quad
    \subcaptionbox{Alopecia}
    {\includegraphics[width=0.30\linewidth, trim={0.5cm 0cm 1.5cm 0.5cm}, 
    clip]{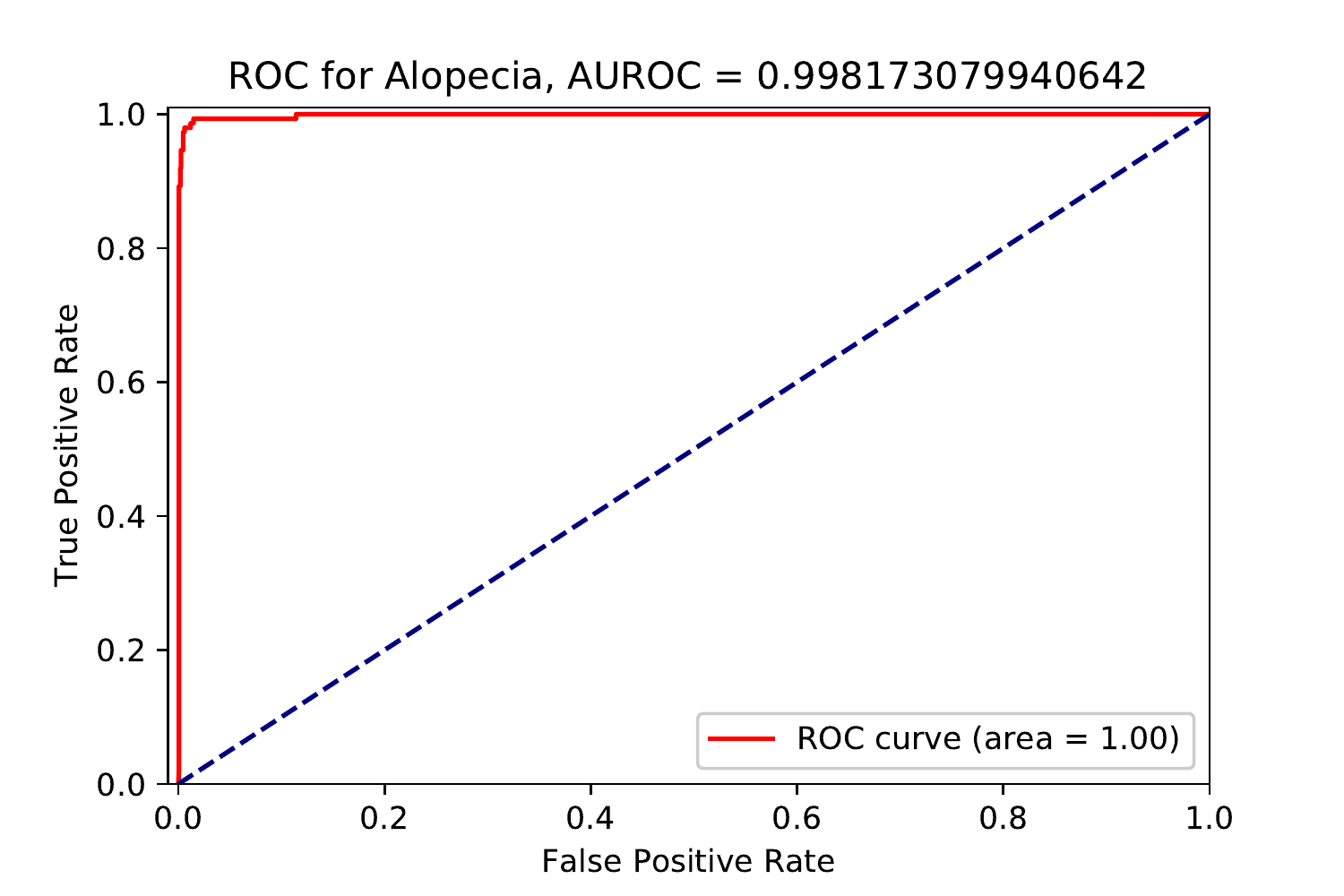}}\quad
    \subcaptionbox{Blister}
    {\includegraphics[width=0.30\linewidth, trim={0.5cm 0cm 1.5cm 0.5cm}, 
    clip]{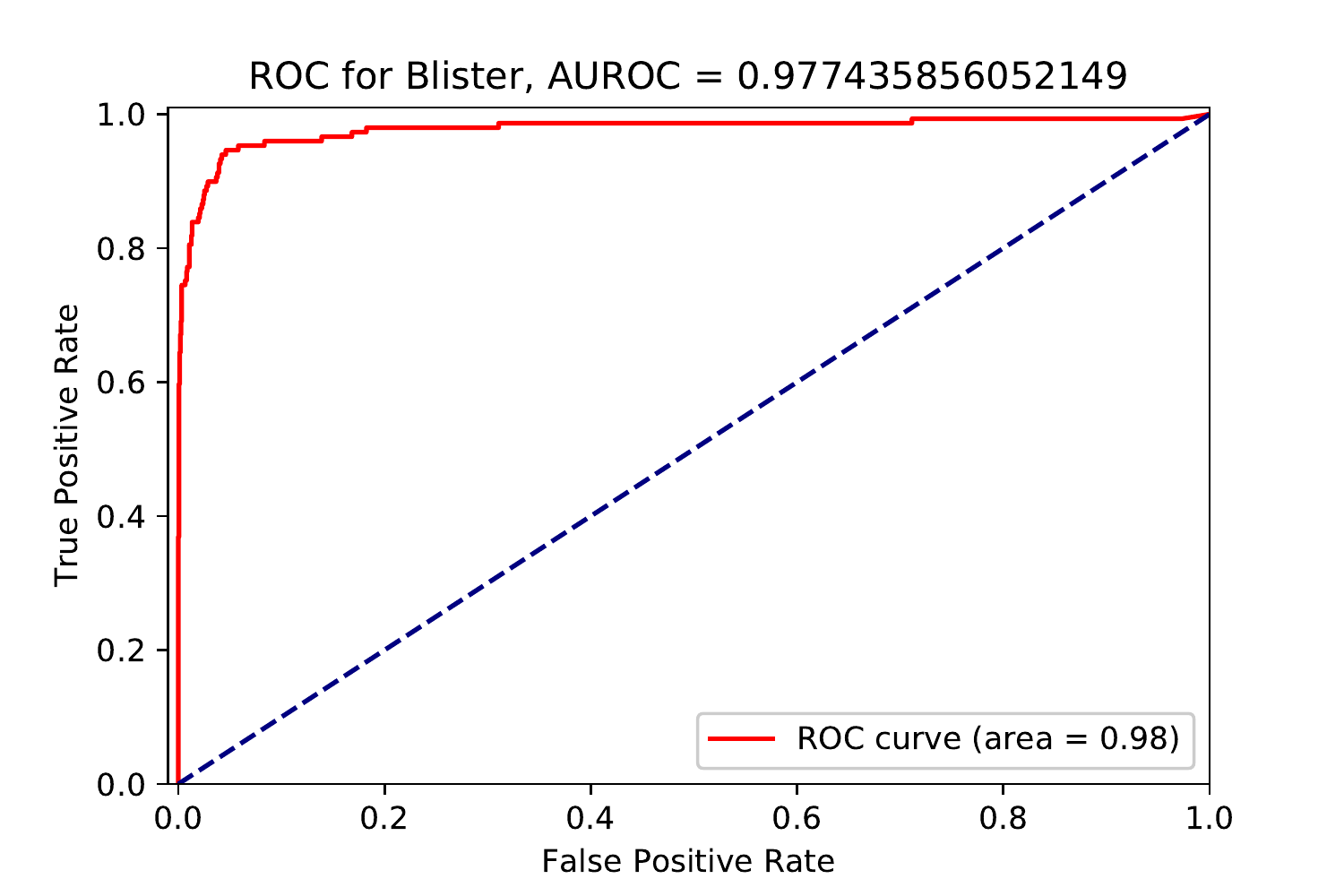}}\quad
    \subcaptionbox{Crust}
    {\includegraphics[width=0.30\linewidth, trim={0.5cm 0cm 1.5cm 0.5cm}, 
    clip]{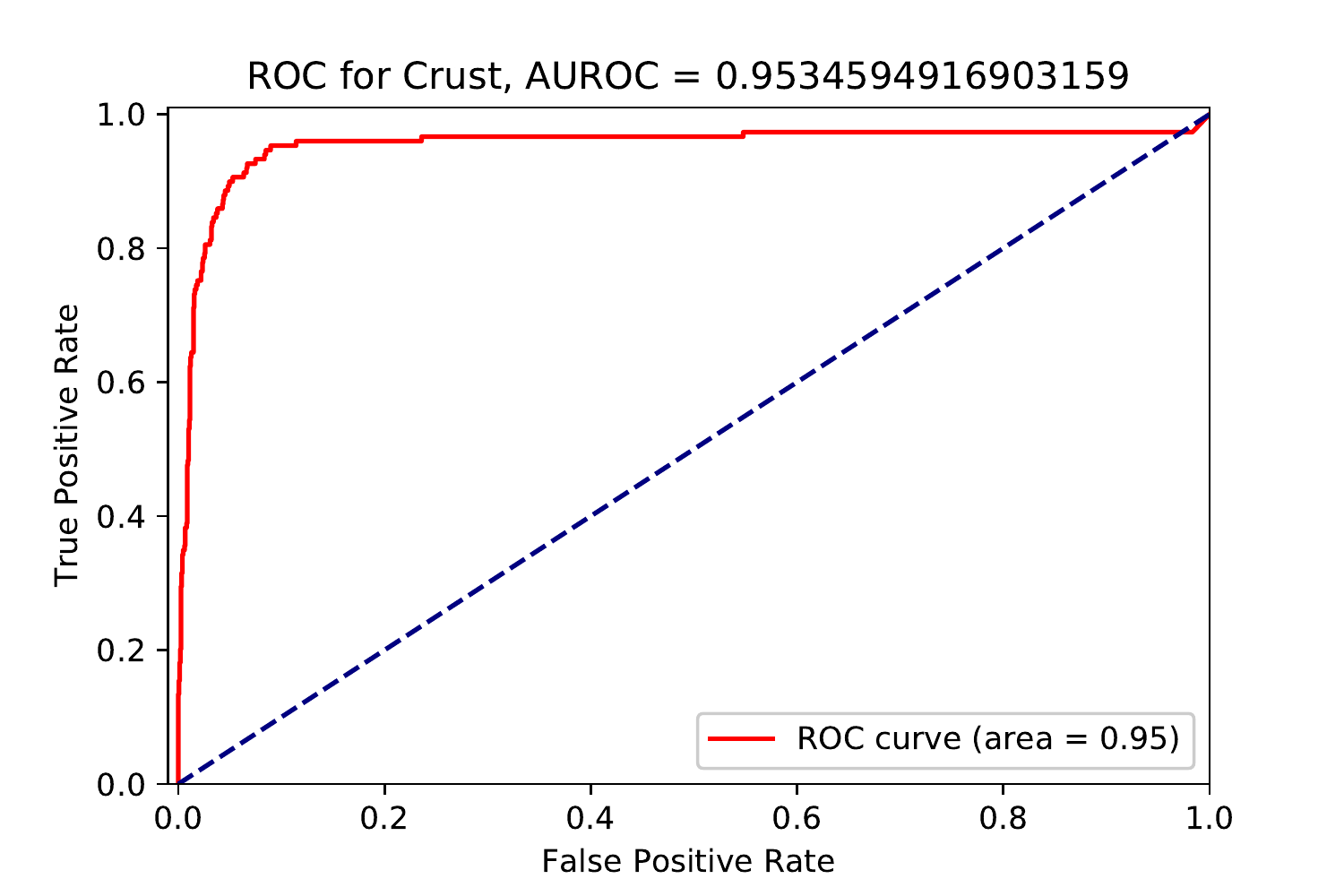}}\quad
    \subcaptionbox{Erythema}
    {\includegraphics[width=0.30\linewidth, trim={0.5cm 0cm 1.5cm 0.5cm}, 
    clip]{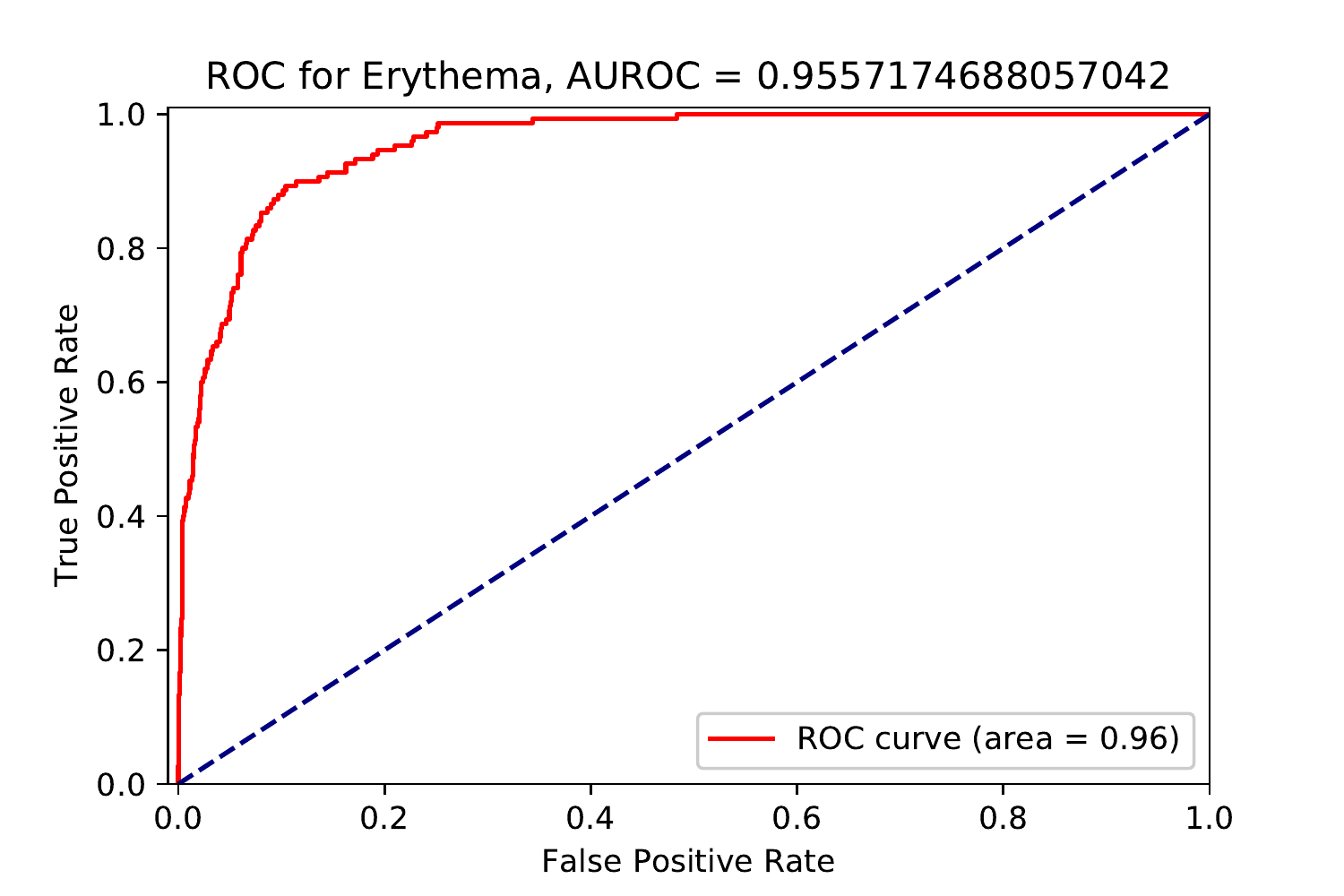}}\quad
    \subcaptionbox{Leukoderma}
    {\includegraphics[width=0.30\linewidth, trim={0.5cm 0cm 1.5cm 0.5cm}, 
    clip]{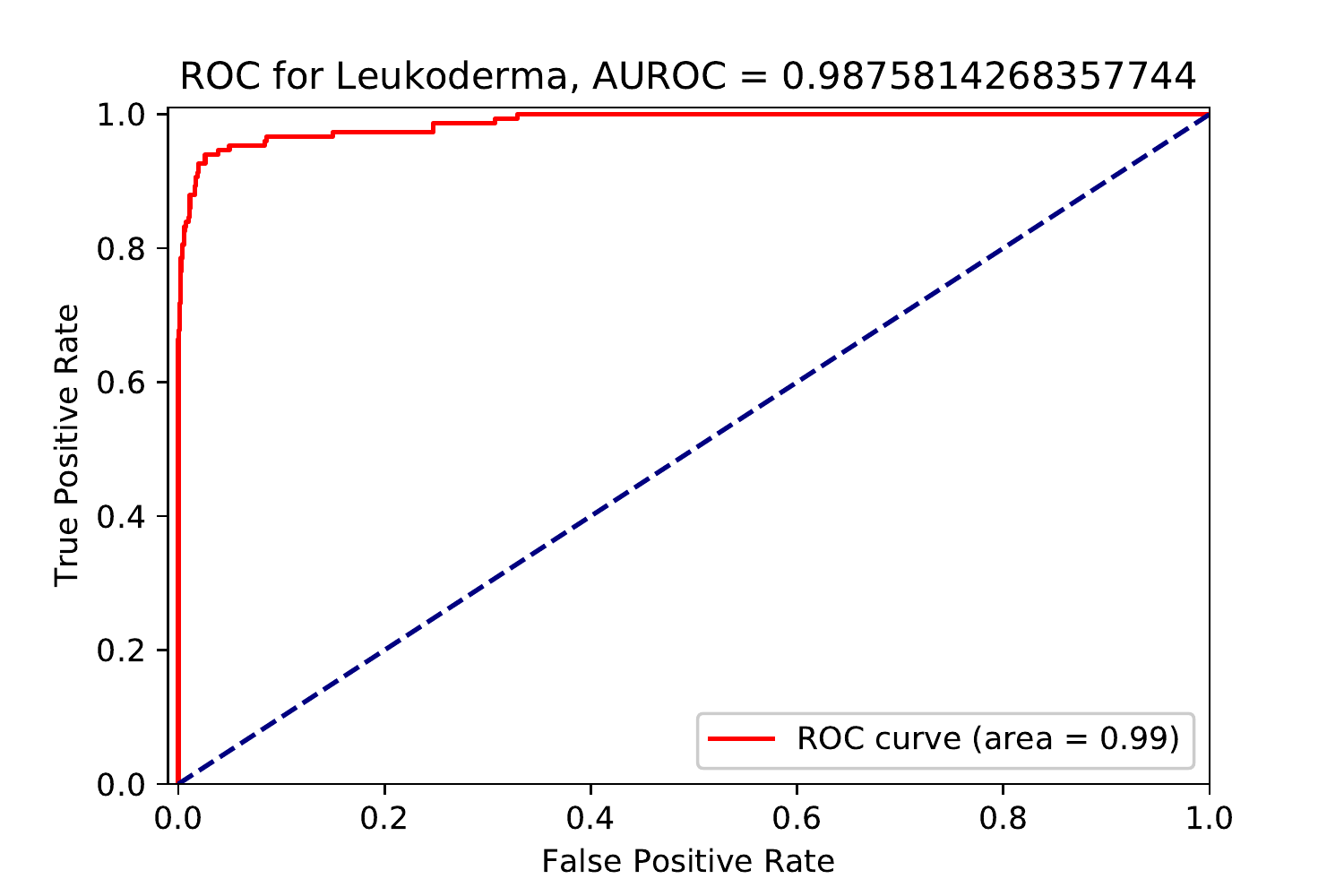}}\quad
    \subcaptionbox{P. Macula}
    {\includegraphics[width=0.30\linewidth, trim={0.5cm 0cm 1.5cm 0.5cm}, 
    clip]{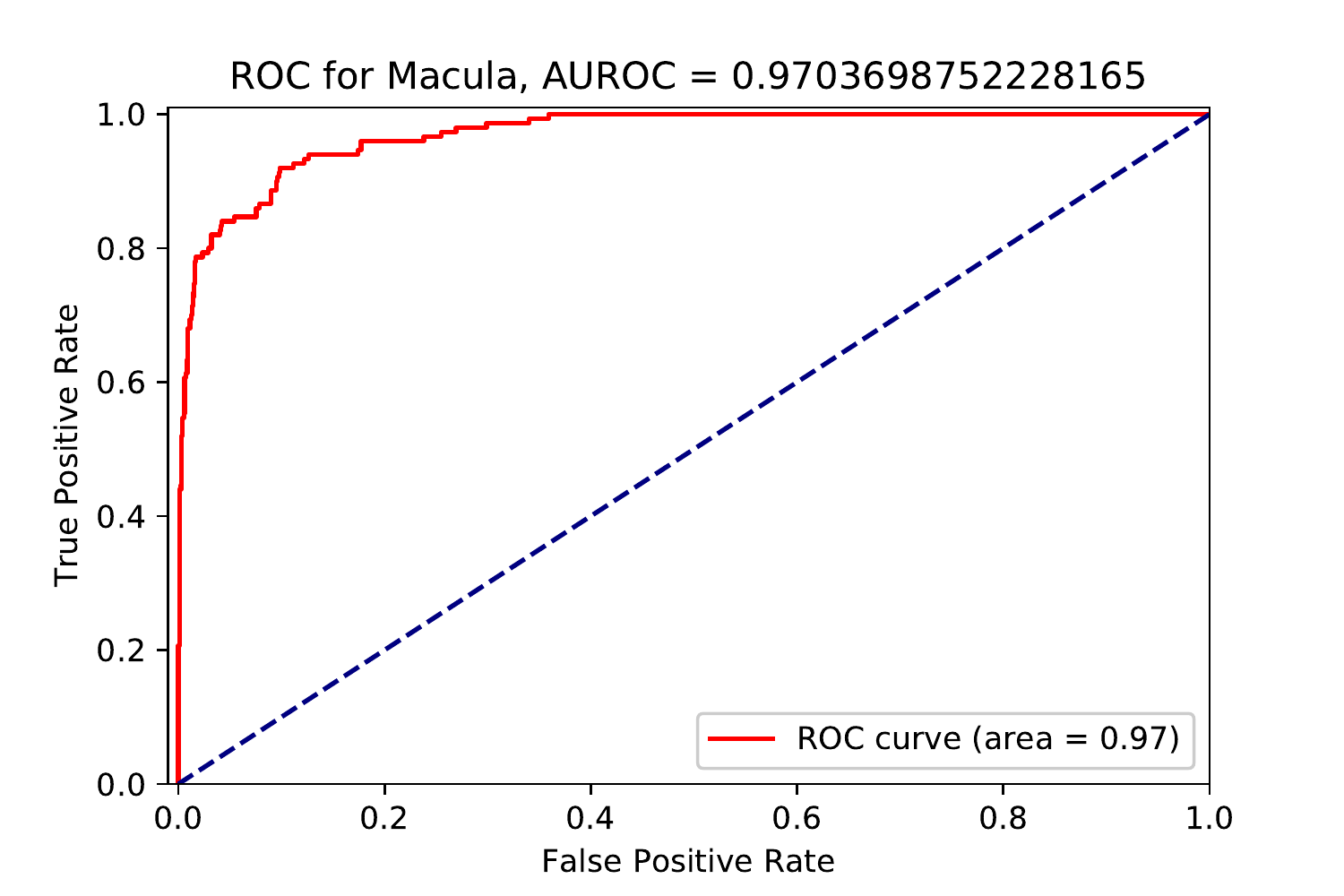}}\quad
    \subcaptionbox{Tumor}
    {\includegraphics[width=0.30\linewidth, trim={0.5cm 0cm 1.5cm 0.5cm}, 
    clip]{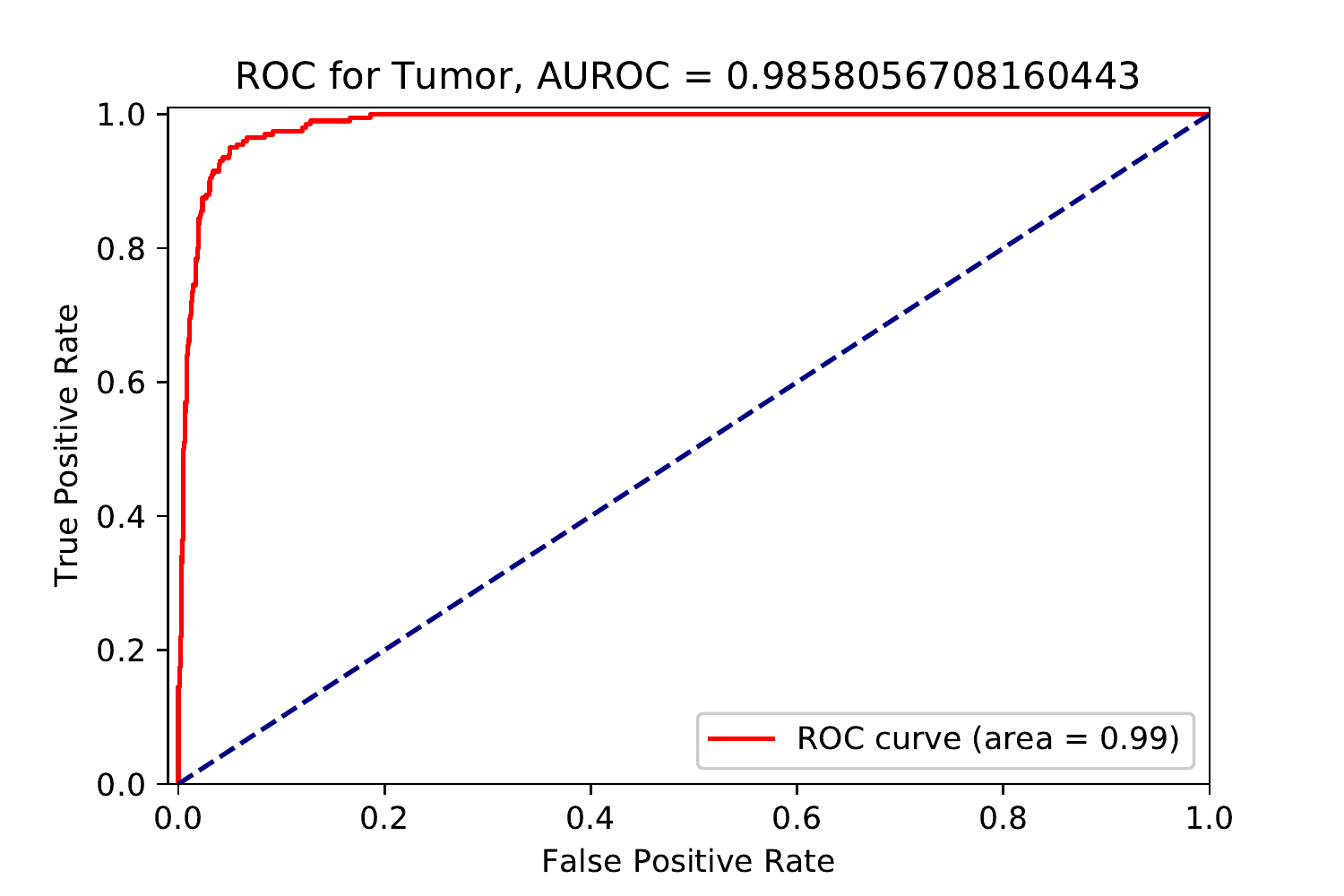}}\quad
    \subcaptionbox{Ulcer}
    {\includegraphics[width=0.30\linewidth, trim={0.5cm 0cm 1.5cm 0.5cm}, 
    clip]{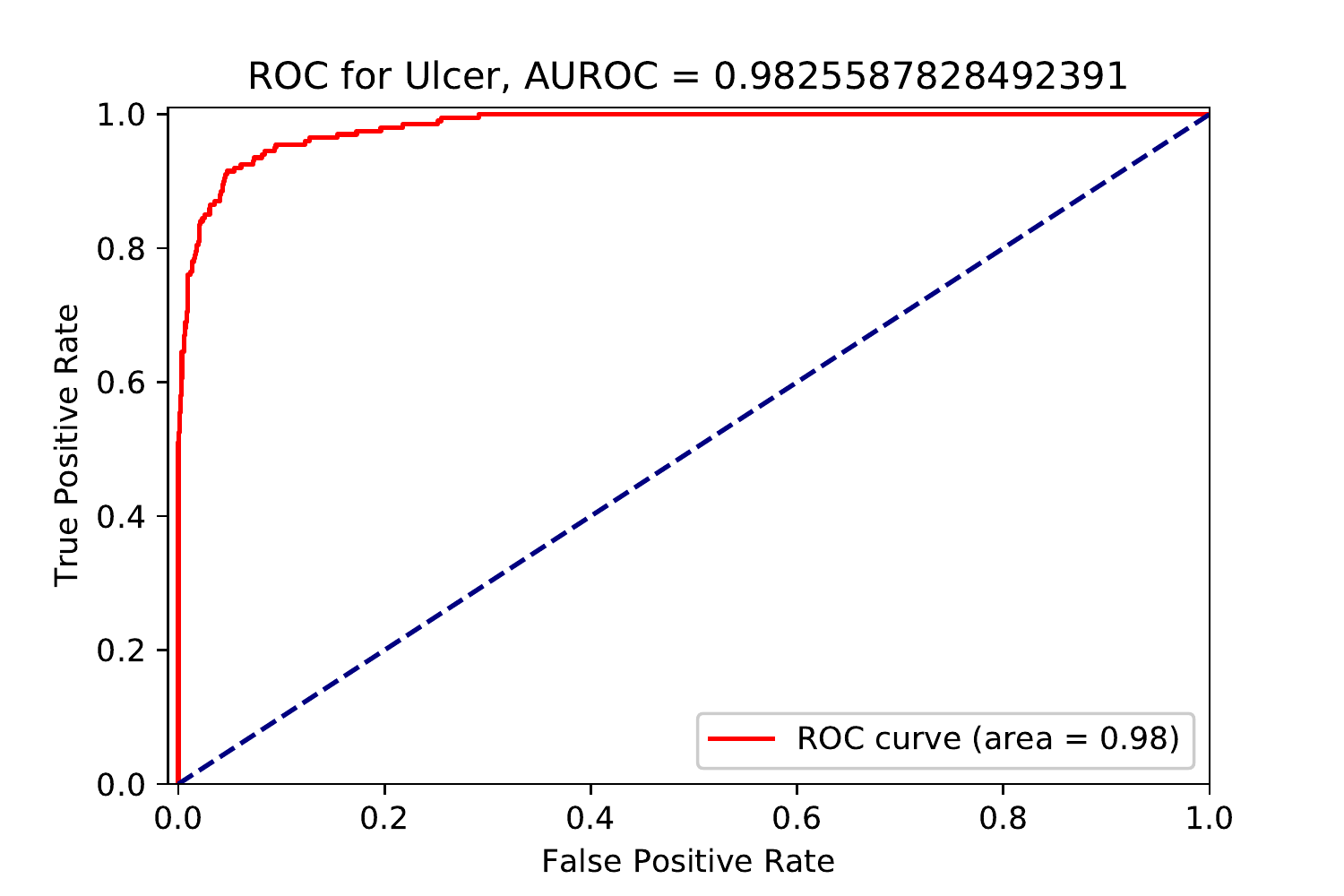}}\quad
    \subcaptionbox{Wheal}
    {\includegraphics[width=0.30\linewidth, trim={0.5cm 0cm 1.5cm 0.5cm}, 
    clip]{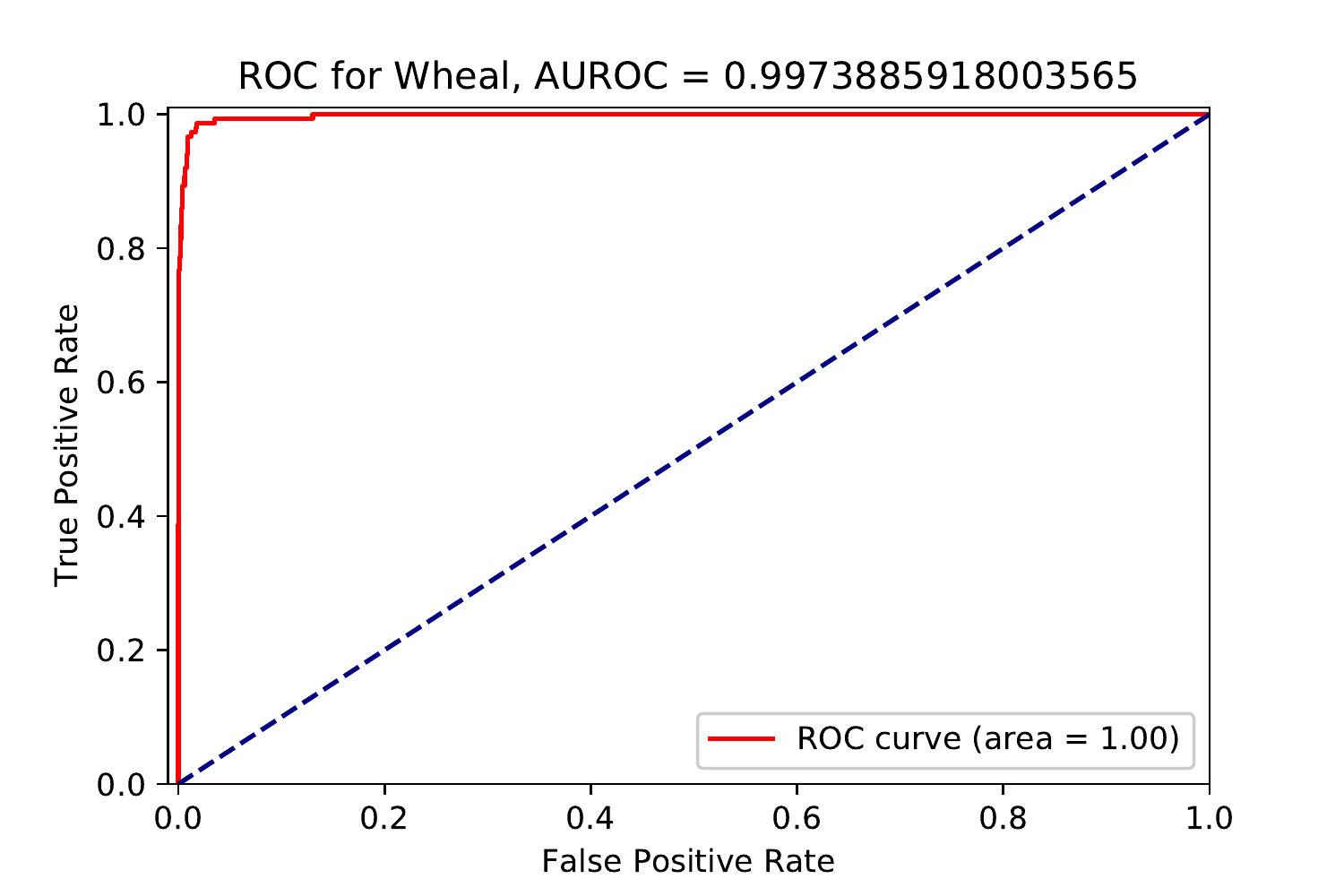}}\quad
    \caption{ROC curves with AUROC score for each class}
    \label{fig:auc_roc}
  \end{minipage}
\end{figure*}

In Sec.~\ref{model-learning}, we trained several ResNet models of 
different sizes to their best fits. Given sufficient epochs, we 
observed their accuracy to be converging to similar scores. All the 
models exhibited aggregate Top-1 accuracy higher than 85\%.
The differences due to architecture sizes were not starkly different.
The accuracy change between smallest and the largest model was 
found to be approximately 3.5\%. The learning stability was evidenced 
by the loss curves (training and validation) in each case, which 
indicated limited possibility for further fit. The results of model 
learning averaged over three trials are shown in 
Table~\ref{tab:exmd-model-learn}. A plot indicating the loss changes 
during ResNet-152 model learning is illustrated in 
Figure~\ref{fig:learning-curves}. The confusion matrix of its 
classification is shown in Figure~\ref{fig:exmd-confusion-R152}.
The receiver operator characteristics (ROC) and area under the curve 
(AUROC) for each class is indicated in Figure~\ref{fig:auc_roc}.

\begin{table}[t]
\caption{Model performance over different architectures}
\label{tab:exmd-model-learn}
\begin{tabular}{lc}
\toprule
    Model & Top-1 Accuracy (in \%) ($\mu\pm\sigma$)\\
\midrule
    ResNet-34   & 86.1 $\pm$ 0.6\\
    ResNet-50   & 86.9 $\pm$ 0.5\\
    ResNet-101  & 89.7 $\pm$ 0.5\\
    ResNet-152  & 89.4 $\pm$ 0.4\\
  \bottomrule
\end{tabular}
\end{table}

\subsection{Effect of lesion variety}
\label{result-sample}
With a stable convergence of different models, we could 
hypothesize that several incorrect classifications arose 
due to the nature of data and incorrect labeling, if any. 
There were certain label pairs which exhibited high degree 
of errors. The ensemble statistics are listed in 
Table~\ref{tab:exmd-toperr}.

\begin{table}[t]
\caption{Skin labels with high rates of confusion}
\label{tab:exmd-toperr}
\begin{tabular}{lc}
\toprule
    Labels & Avg. Erroneous predictions ($\mu\pm\sigma$)\\
\midrule
    \textit{Ulcer} and \textit{Tumor} & $31.75\pm6.28$\\
    \textit{P. Macula} and \textit{Erythema} & $27.25\pm3.50$\\
    \textit{Erythema} and \textit{Wheal} & $19.25\pm2.75$\\
    \textit{Crust} and \textit{Ulcer} & $16.00\pm3.46$\\
  \bottomrule
\end{tabular}
\end{table}

\begin{figure*}
\begin{minipage}[t]{.49\linewidth}
    \begin{center}
    \includegraphics[width=\linewidth, trim={5cm 2.5cm 5cm 1cm}, 
    clip]{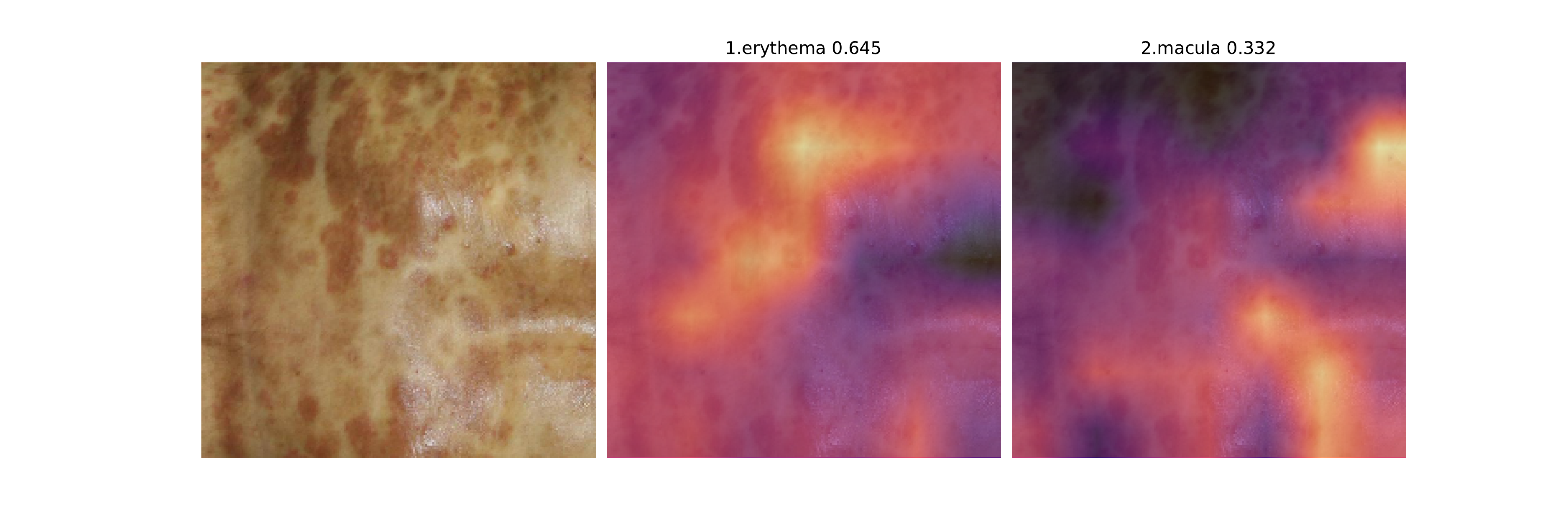}
    \end{center}
    \caption{\textit{P. Macula} misclassified as \textit{Erythema}}
    \label{fig:macula-erythema}
\end{minipage}
\hfill
\begin{minipage}[t]{.49\linewidth}
    \begin{center}
        \includegraphics[width=\linewidth, trim={5cm 2.5cm 5cm 1cm}, 
        clip]{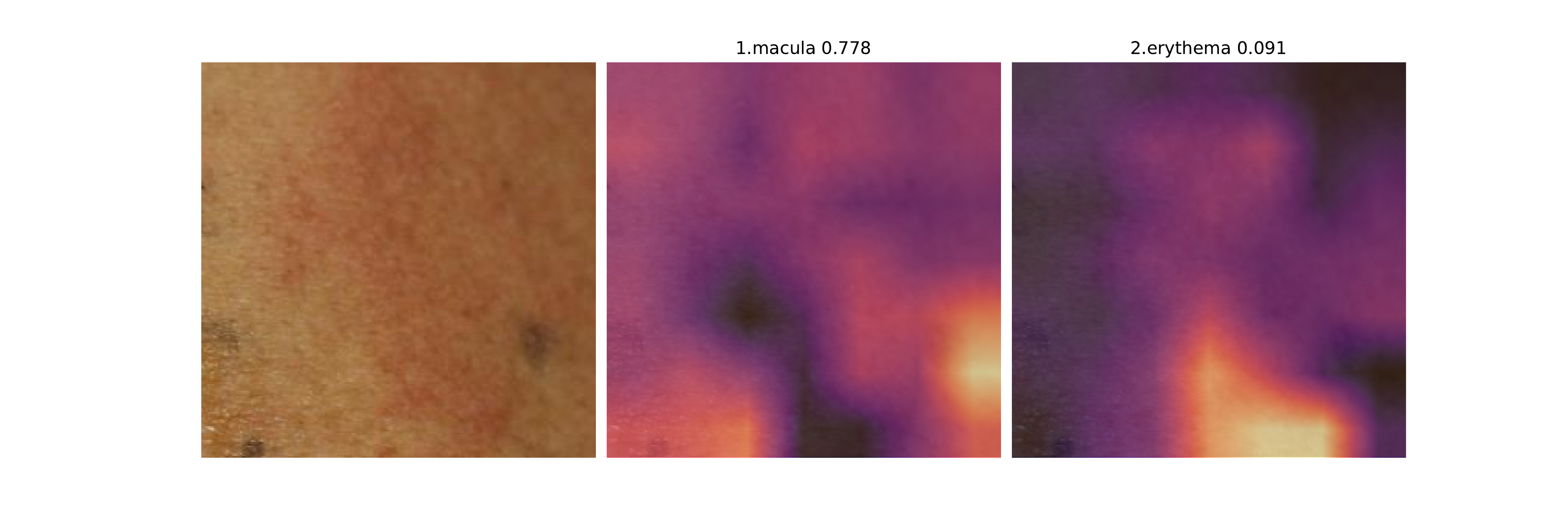}
    \end{center}
    \caption{\textit{Erythema} misclassified as \textit{P. Macula}}
    \label{fig:erythema-macula}
\end{minipage}

\begin{minipage}[t]{.49\linewidth}
    \begin{center}
    \includegraphics[width=\linewidth, trim={5cm 2.5cm 5cm 0cm}, 
    clip]{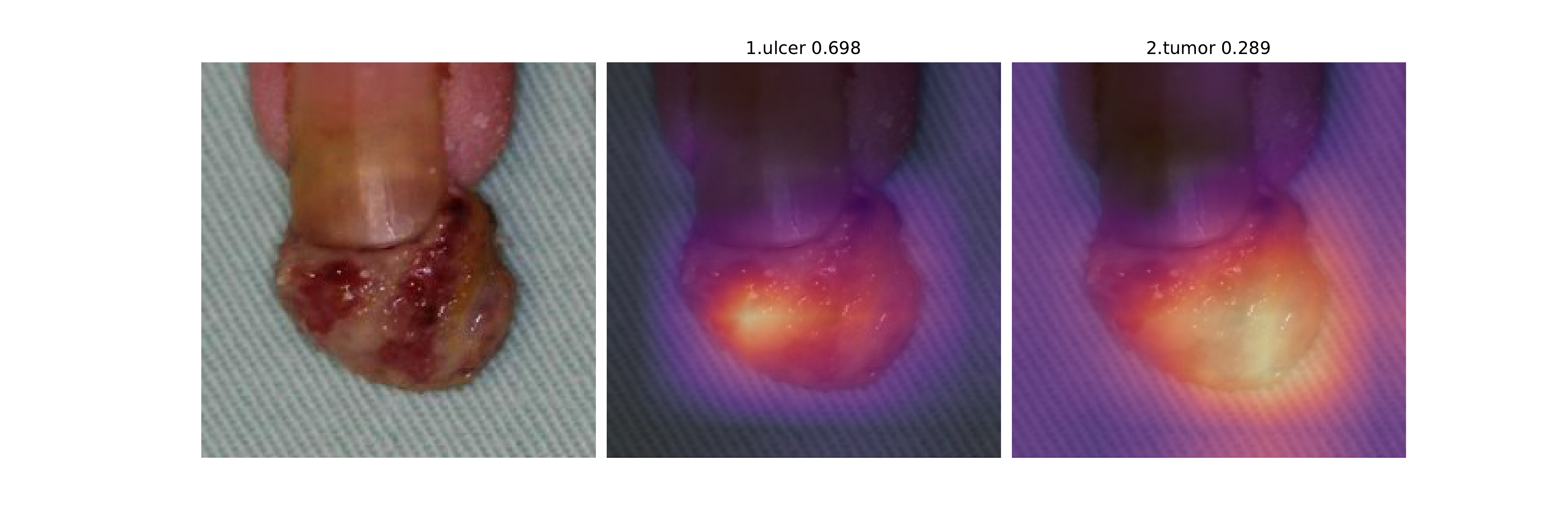}
    \end{center}
    \caption{\textit{Tumor} misclassified as \textit{Ulcer}}
    \label{fig:tum-ulc}
\end{minipage}
\hfill
\begin{minipage}[t]{.49\linewidth}
    \begin{center}
    \includegraphics[width=\linewidth, trim={5cm 2.5cm 5cm 0cm}, 
    clip]{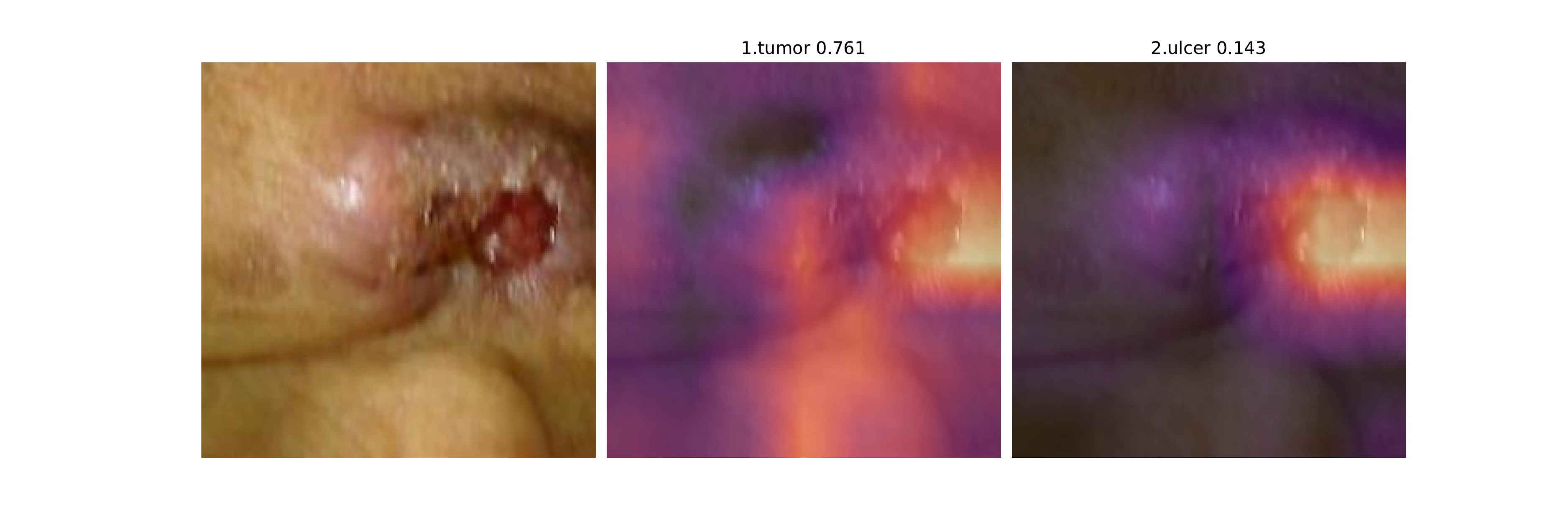}
    \end{center}
    \caption{\textit{Ulcer} misclassified as \textit{Tumor}}
    \label{fig:ulc-tum}
\end{minipage}
\end{figure*}

We investigated some erroneous predictions using 
GradCAM~\cite{selvaraju2017}. An example of \textit{Hyperplasic 
Pigmented Macula} is exhibited in Figure~\ref{fig:macula-erythema}. 
The classifier had a shape bias to detect it as \textit{Erythema}, 
since a majority of the latter present themselves in contrasting 
patches. An example of \textit{Erythema} incorrectly classified 
as \textit{P. Macula} is shown in Figure~\ref{fig:erythema-macula}. 
The presence of hyperpigmented spots in the periphery drew the 
classifier to believe it to be of the wrong class. Texture bias 
in vision models discussed by Geirhos et al. was evident in the 
examples of \textit{Ulcer} misclassified as \textit{Tumor}, and 
vice versa (Figures~\ref{fig:tum-ulc} and 
\ref{fig:ulc-tum})~\cite{geirhos2018}. The absence of any 
identifying texture led to higher confusion while categorizing 
between \textit{Erythema} and \textit{Wheal}.



\subsection{Effect of adversarial artifacts}
\label{result-artifacts}
\begin{figure}[t]
    \includegraphics[width=\linewidth,trim={2.5cm 1cm 2.5cm 0cm}, 
    clip]{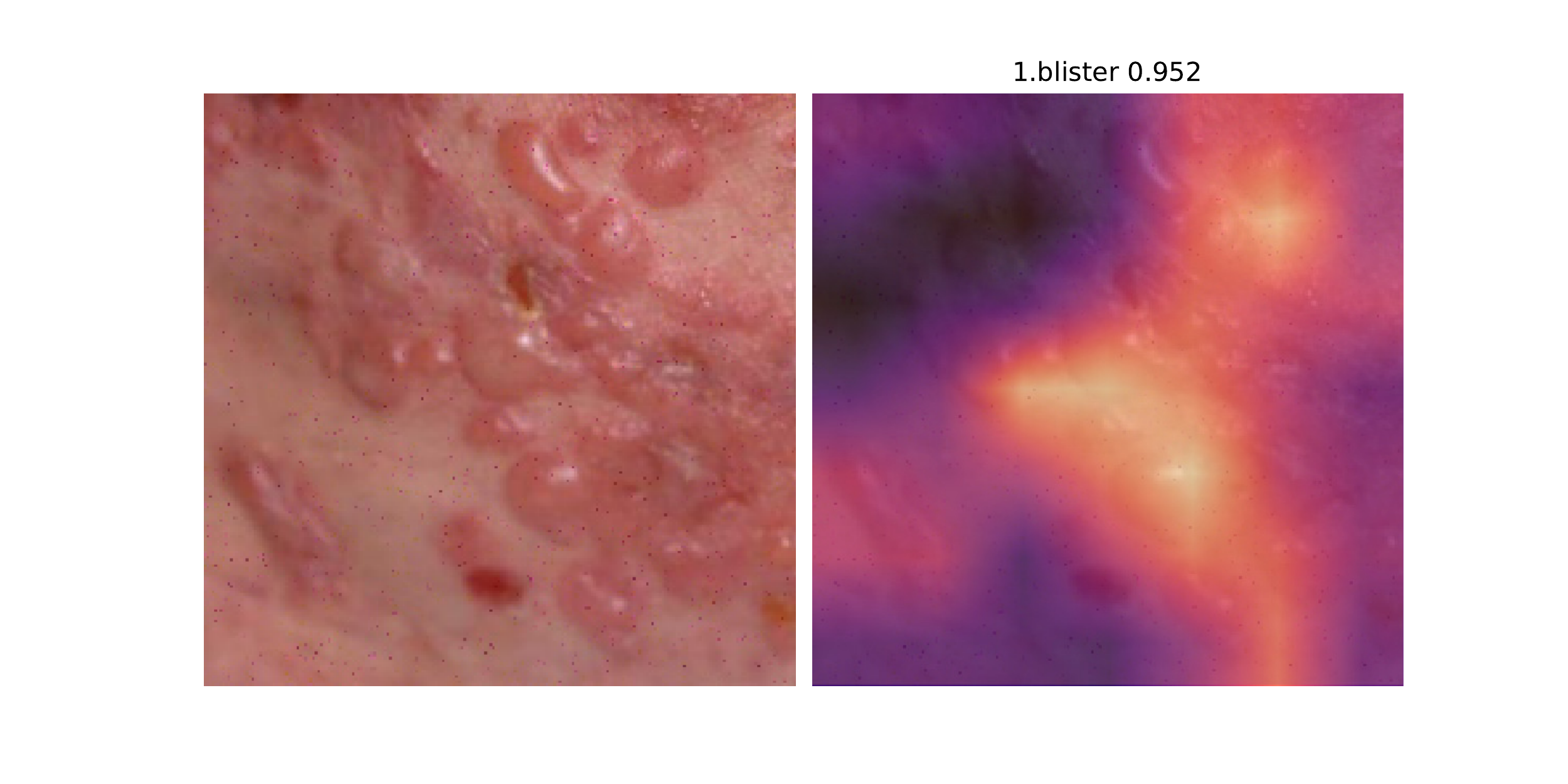}
    \caption{Prediction of \textit{Blister} to its true class}
    \label{fig:sp_correct}
\end{figure}

\begin{figure}[t]
    \includegraphics[width=\linewidth,trim={3cm 1cm 3cm 0cm}, 
    clip]{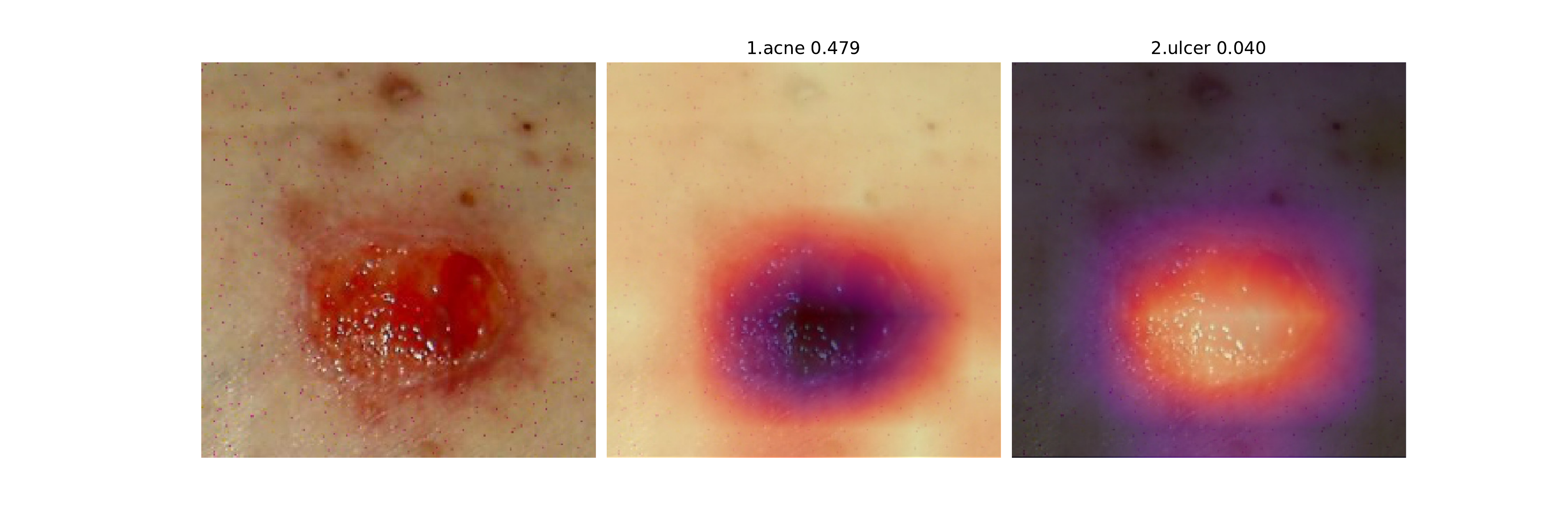}
    \caption{Prediction change (\textit{Ulcer} to \textit{Acne})
    in the presence of S\&P noise.}
    \label{fig:sp_incorrect}
\end{figure}

In Sec.~\ref{adversarial-test}, we described the means of creating 
adversarial examples to measure the robustness of model against 
shot noise and motion blur. Using ResNet-50 trained on uncorrupted 
data, we measured the classification performance of the model on 
test set having up to 5\% corrupted pixels. The Top-1 accuracy under 
this setting was observed to be 68\%. Figure~\ref{fig:sp_correct} 
shows correct prediction for the sample shown in Figure~\ref{fig:sp_corruption} 
despite the presence of S\&P noise. However, many otherwise obvious 
cases show sharp prediction changes, such as Figure~\ref{fig:sp_incorrect} 
for a sample of \textit{Ulcer}. We also observed emergence of a bias 
for the classifier towards a few labels, such as \textit{Acne}, 
\textit{Blister} and \textit{Wheal}.

\begin{figure}[t]
    \includegraphics[width=\linewidth,trim={2.5cm 1cm 2.5cm 0cm}, 
    clip]{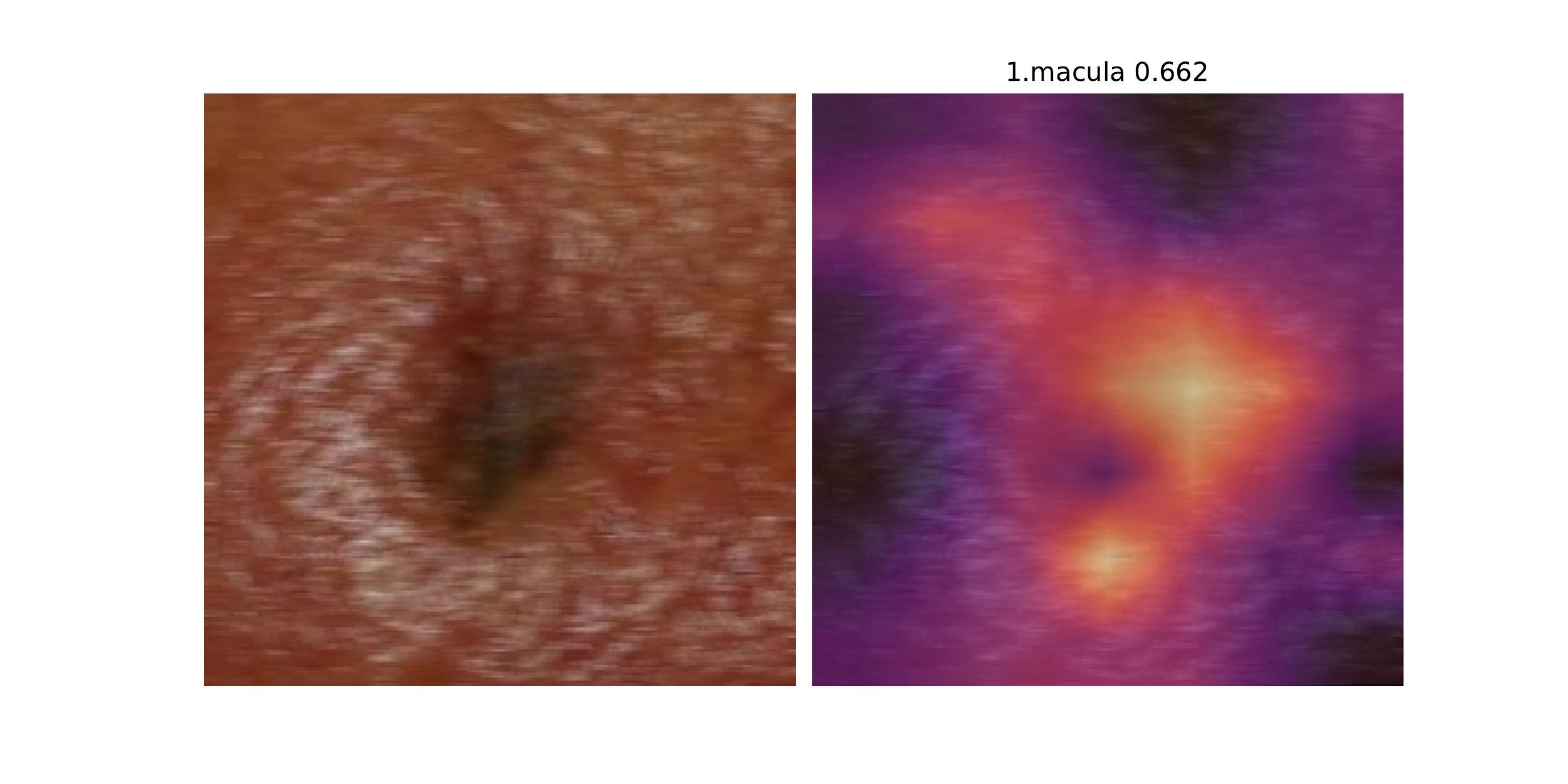}
    \caption{Prediction of \textit{P. Macula} to its correct class}
    \label{fig:blur_correct}
\end{figure}

\begin{figure}[t]
    \includegraphics[width=\linewidth]{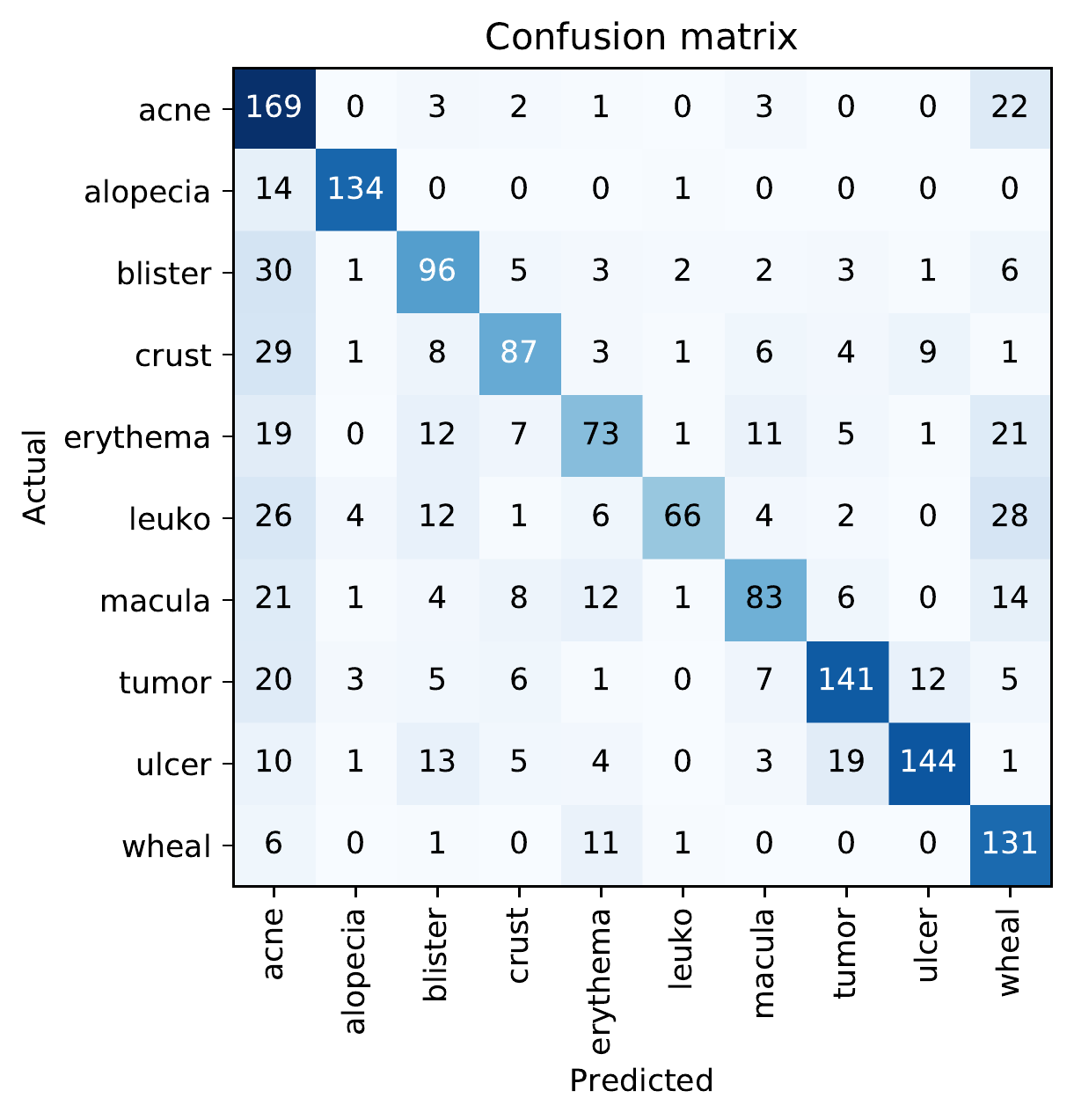}
    \caption{Results of adversarial testing in presence of 
    Salt \& pepper noise.}
    \label{fig:exmd-confusion-val-sp}
\end{figure}

\begin{figure}[t]
    \includegraphics[width=\linewidth]{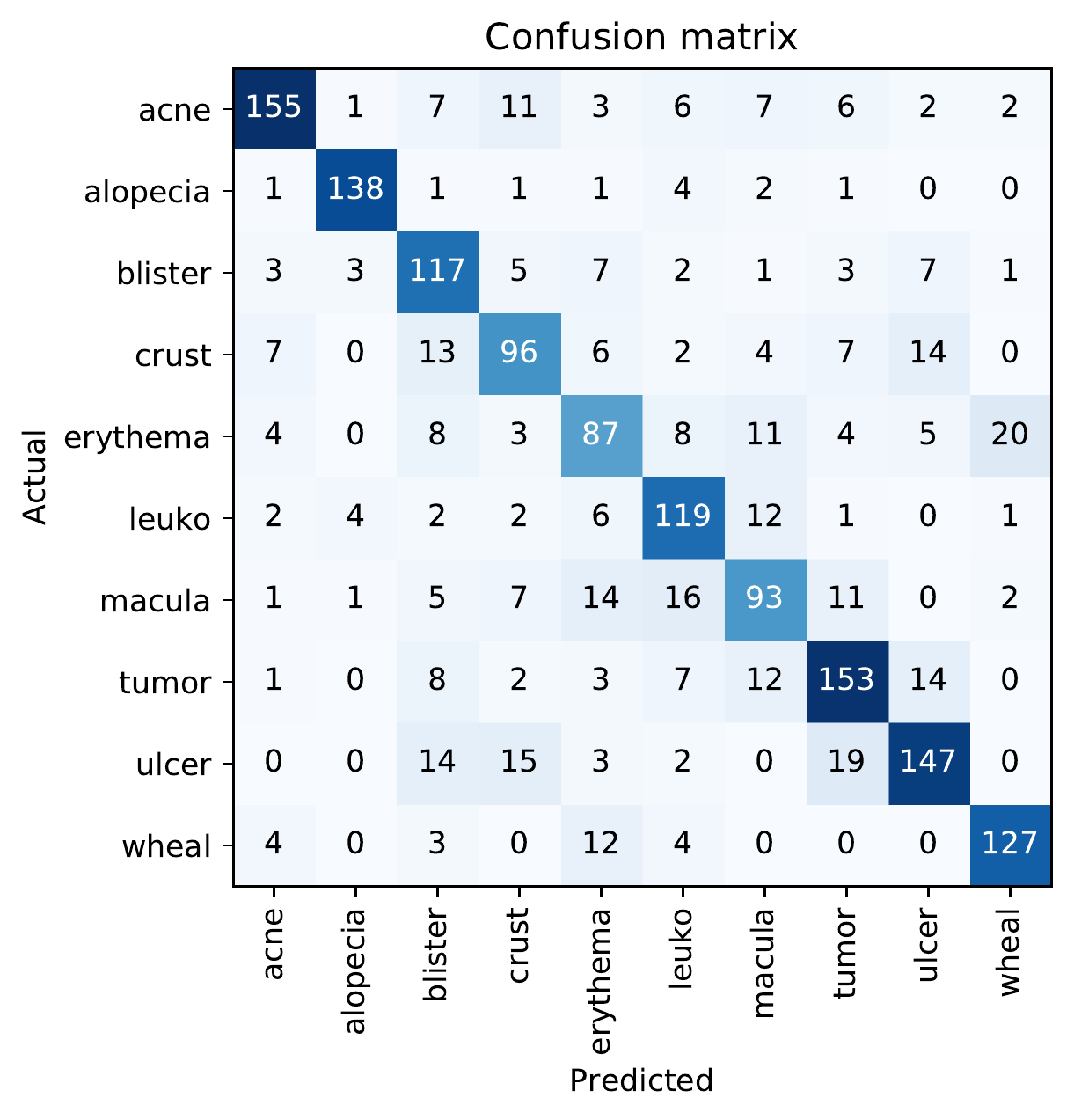}
    \caption{Results of adversarial testing in presence of blur}
    \label{fig:exmd-confusion-val-blur}
\end{figure}

In the presence of motion blur, the performance of classifier was 
marginally better at 74.85\%. Figure~\ref{fig:blur_correct} shows 
the previously illustrated example (Fig.~\ref{fig:blur_corruption}) 
being correctly identified to its true class label. Although bias 
did emerge due to introduction of the blur, it was not as severe 
as seen in the presence of shot noise. In both cases however, we 
observed the classification quality change significantly which 
is illustrated in the confusion matrices shown in 
Figures~\ref{fig:exmd-confusion-val-sp} and 
\ref{fig:exmd-confusion-val-blur}. 

We also trained our model on shot noise corrupted images and 
evaluated the performance on both clean and corrupted test sets. 
The Top-1 accuracy measured in these cases were 75.6\% and 80.92\% 
respectively. We observed that the confusion in classes changed 
significantly when the model was trained on clean data and tested 
on corrupted images. The pattern of original class confusions 
were retained to some degree on both clean and corrupted test set 
when the model was trained in the presence of shot noise. 
Table~\ref{tab:exmd-adversrial} gives a brief summary of the 
adversarial experiments conducted.

\begin{table}[t]
\caption{Summary of adversarial tests}
\label{tab:exmd-adversrial}
\begin{tabular}{llc}
\toprule
    Train Status & Test Status & Top-1 Acc. (in \%)\\
\midrule
    Clean w/o Noise & N.A.              & 86.90\\
    Clean w/o Noise & S\&P Corrupted    & 68.31\\
    Clean w/o Noise & Blurred           & 74.85\\
    S\&P Corrupted  & Clean w/o Noise   & 75.60\\
    S\&P Corrupted  & S\&P Corrupted    & 80.92\\
  \bottomrule
\end{tabular}
\end{table}

\subsection{Effect of distribution shift}
\label{result-distribution-shift}

In real world scenarios, any dermatological classifier is expected 
to work on diverse set of input images. A supplied image may be 
different from the kind of images the model is trained on. Moreover,  
input images may also belong to classes which the classifier was 
not designed for. In the cases of distribution shift, and out of 
distribution samples, models are required to behave predictably.

\begin{figure}[t]
    \includegraphics[width=\linewidth]{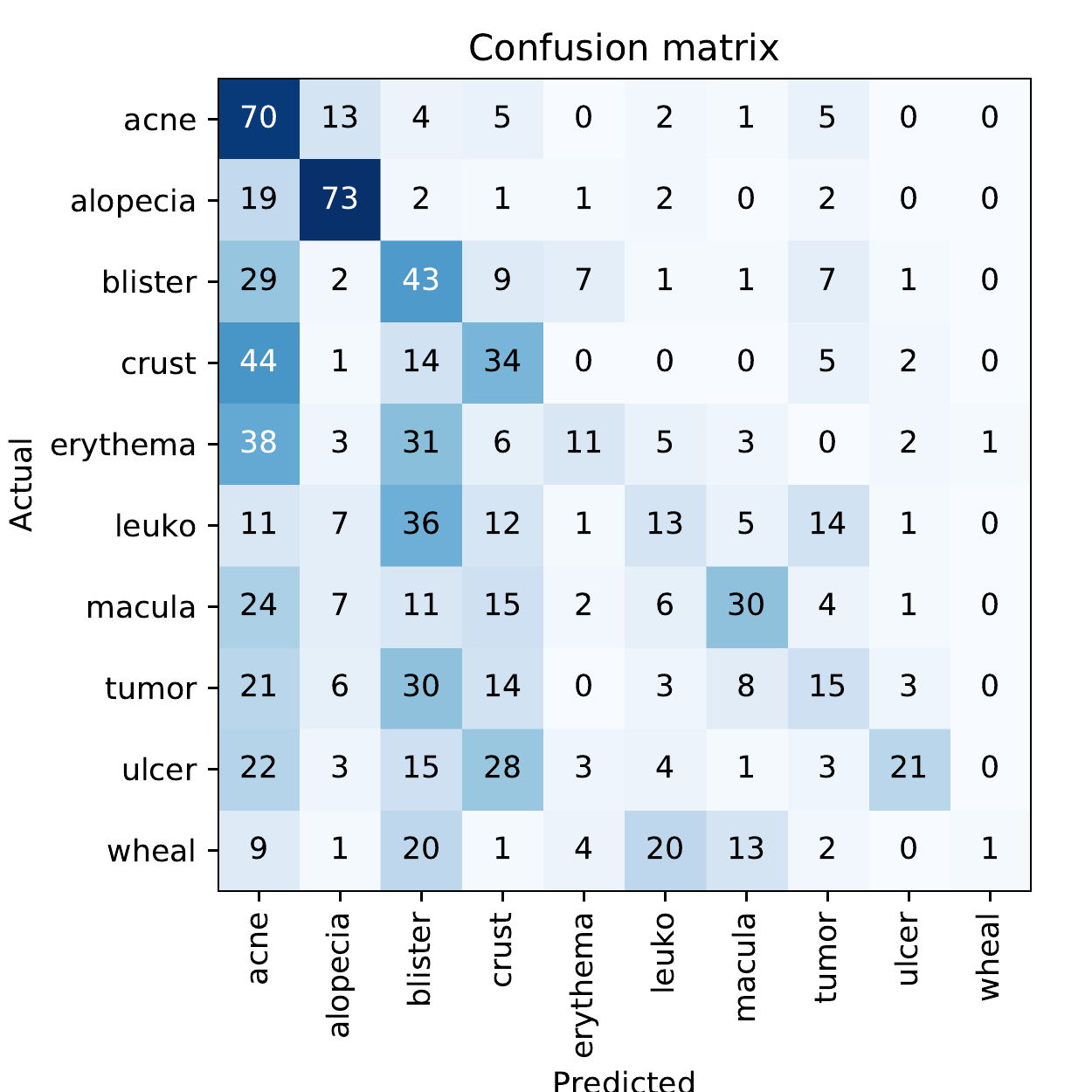}
    \caption{Confusion matrix for inference on samples from 
    select SD-198 classes grouped for out of distribution 
    study}
    \label{fig:ood-test}
\end{figure}

As described in Sec.~\ref{data}, we grouped matching classes of 
SD-198 dataset to our experimental specification for creating a 
composite test set with a hundred samples in each class (Only 
exception being \textit{Wheal} which had 71 samples). We trained 
a ResNet-50 model to optimal accuracy on our dataset, and performed 
inference on the test set to evaluate predictions. Generalization 
was seen to be poor with only 32\% Top-1 aggregate accuracy. Only 
\textit{Acne} and \textit{Alopecia} fared properly with 70\% and 
73\% accuracy respectively. Classifier bias was seen to favor 
\textit{Acne} and \textit{Blister} dominantly over others. 
Figure~\ref{fig:ood-test} shows a confusion matrix for the 
classifier performance on this test set.

\section{Discussion}
In Sec.~\ref{result-learning}, we demonstrated means to perform a 
good model fit on homogeneous skin disease images. From the computer 
vision frame of reference, although background homogeneity may be an 
excellent option, it poses several challenges in dermatological 
classification. Despite of robust training, a sufficiently large 
gap exists between our results and ideal, error-free performance. 
We ascribe it predominantly to the nature of the data itself. 
The visual attributes of skin diseases are not discrete by nature. 
Many diseases belong to wide spectrum and families of abnormalities 
which make their detection and grading difficult. Several of these 
disease classes are also chronologically related. Instances of 
\textit{Acne} and \textit{Blister} can eventually lead to either 
formation of \textit{Ulcers} or \textit{Crusts}. Different pathways 
exist for lesion progression, increasing the complexity. 
Although it could be obvious for a trained clinician, intermediate 
transitional states between classes could also be a difficult 
challenge in machine diagnosis. Skin diseases also rarely occur 
in isolation. Conditions such as \textit{Erythema} or 
\textit{Hyperpigmentation} are concurrent in presentation 
with several other diseases. Dermatological classifiers can be biased 
in detecting one over the other. In the absence of valuable patient 
history, the information presented is only skin-deep. 

Unlike radiological images, skin disease images present very few 
landmarks. There are only a handful of diseases with strong visual 
markers or patterns to aid classification. This issue is compounded 
by the large spectrum of skin tones in the patient pool. Lesion 
contrast and color are variable attributes when considering racial 
differences. Although some remedial measures have been discussed 
previously, contrast and illumination correction may not be 
sufficient~\cite{mishra2019interpreting}. Our models were trained on 
a single racial type. With the introduction of more racial cohorts, 
we strongly believe the class confusions will aggravate. We 
demonstrated this effect to some extent in 
Sec.~\ref{result-distribution-shift}, where model testing on 
composite similar classes dramatically reduced the efficiency of 
the classifier. Irregularities tied to the nature of data may 
not be alleviated in the near future. 

Adversarial attacks can cause wrong categorizations with seemingly 
unchanged input images. We presented adversarial examples via 
shot noise \& simulated motion blur which degraded the performance of 
classification. User submitted images typically contain some degree 
of such artifacts. Dermatological classifiers have not yet demonstrated 
confidence in overcoming them. We plan to study the effects of 
adversarial perturbations in skin images to design better 
classification models in the future. This might also help in 
reducing vulnerabilities from malicious attacks in real world 
systems.

In light of these deficiencies, human expertise cannot be be 
removed from the clinical workflow. The margin of error in healthcare 
is slim. In addition to better and varied skin data, we require 
clinicians in correctly labeling skin conditions and verifying 
diagnostic outcomes. From our experience, computer vision techniques 
have been useful in detecting mislabeled samples or the presence of 
a novel category in our experimental data. But a trained practitioner 
is required eventually to appraise this information. At present, even if 
deep learning based classifiers cannot replace the human expertise, they 
can be valuable physician aids in screening patients.

\section{Conclusion}
In this paper, we demonstrated that several skin diseases can be 
identified from user submitted images with deep learning based 
classifiers. We showed that given sufficient training, the accuracy 
levels become architecture agnostic. There exists a significant gap 
between error-free detection and the peak performance achieved by 
contemporary methods. This gap may not be bridged easily since it 
manifests from the nature of skin disease presentation. We also 
showed that the performance dipped by at least 10\% in non-ideal 
conditions such as noise, blur and distribution shift, which are 
reasonable scenarios in any field trial. We emphasize the role of 
trained practitioners in conjunction with these methods to improve 
the quality of dermatological services. There is a long road yet 
to be travelled to achieve unassisted, reliable diagnosis of skin 
conditions. 

\bibliographystyle{ACM-Reference-Format}
\bibliography{sample-base}


\begin{thebibliography}{27}


\ifx \showCODEN    \undefined \def \showCODEN     #1{\unskip}     \fi
\ifx \showDOI      \undefined \def \showDOI       #1{#1}\fi
\ifx \showISBNx    \undefined \def \showISBNx     #1{\unskip}     \fi
\ifx \showISBNxiii \undefined \def \showISBNxiii  #1{\unskip}     \fi
\ifx \showISSN     \undefined \def \showISSN      #1{\unskip}     \fi
\ifx \showLCCN     \undefined \def \showLCCN      #1{\unskip}     \fi
\ifx \shownote     \undefined \def \shownote      #1{#1}          \fi
\ifx \showarticletitle \undefined \def \showarticletitle #1{#1}   \fi
\ifx \showURL      \undefined \def \showURL       {\relax}        \fi
\providecommand\bibfield[2]{#2}
\providecommand\bibinfo[2]{#2}
\providecommand\natexlab[1]{#1}
\providecommand\showeprint[2][]{arXiv:#2}

\bibitem[\protect\citeauthoryear{Allen}{Allen}{2019}]%
        {allen2019role}
\bibfield{author}{\bibinfo{person}{Bibb Allen}.}
  \bibinfo{year}{2019}\natexlab{}.
\newblock \showarticletitle{The role of the FDA in ensuring the safety and
  efficacy of artificial intelligence software and devices}.
\newblock \bibinfo{journal}{\emph{Journal of the American College of
  Radiology}} \bibinfo{volume}{16}, \bibinfo{number}{2} (\bibinfo{year}{2019}),
  \bibinfo{pages}{208--210}.
\newblock


\bibitem[\protect\citeauthoryear{Codella, Nguyen, Pankanti, Gutman, Helba,
  Halpern, and Smith}{Codella et~al\mbox{.}}{2017}]%
        {codella2017}
\bibfield{author}{\bibinfo{person}{Noel~CF Codella}, \bibinfo{person}{Q-B
  Nguyen}, \bibinfo{person}{Sharath Pankanti}, \bibinfo{person}{David~A
  Gutman}, \bibinfo{person}{Brian Helba}, \bibinfo{person}{Allan~C Halpern},
  {and} \bibinfo{person}{John~R Smith}.} \bibinfo{year}{2017}\natexlab{}.
\newblock \showarticletitle{Deep learning ensembles for melanoma recognition in
  dermoscopy images}.
\newblock \bibinfo{journal}{\emph{IBM Journal of Research and Development}}
  \bibinfo{volume}{61}, \bibinfo{number}{4/5} (\bibinfo{year}{2017}),
  \bibinfo{pages}{5--1}.
\newblock


\bibitem[\protect\citeauthoryear{Dekio, Hanada, Chinuki, Akaki, Kitani,
  Shiraishi, Kaneko, Furumura, and Morita}{Dekio et~al\mbox{.}}{2010}]%
        {dekio2010}
\bibfield{author}{\bibinfo{person}{Itaru Dekio}, \bibinfo{person}{Eisuke
  Hanada}, \bibinfo{person}{Yuko Chinuki}, \bibinfo{person}{Tatsuya Akaki},
  \bibinfo{person}{Mitsuhiro Kitani}, \bibinfo{person}{Yuko Shiraishi},
  \bibinfo{person}{Sakae Kaneko}, \bibinfo{person}{Minao Furumura}, {and}
  \bibinfo{person}{Eishin Morita}.} \bibinfo{year}{2010}\natexlab{}.
\newblock \showarticletitle{Usefulness and economic evaluation of ADSL-based
  live interactive teledermatology in areas with shortage of dermatologists}.
\newblock \bibinfo{journal}{\emph{International journal of dermatology}}
  \bibinfo{volume}{49}, \bibinfo{number}{11} (\bibinfo{year}{2010}),
  \bibinfo{pages}{1272--1275}.
\newblock


\bibitem[\protect\citeauthoryear{Esteva, Kuprel, Novoa, Ko, Swetter, Blau, and
  Thrun}{Esteva et~al\mbox{.}}{2017}]%
        {esteva2017}
\bibfield{author}{\bibinfo{person}{Andre Esteva}, \bibinfo{person}{Brett
  Kuprel}, \bibinfo{person}{Roberto~A Novoa}, \bibinfo{person}{Justin Ko},
  \bibinfo{person}{Susan~M Swetter}, \bibinfo{person}{Helen~M Blau}, {and}
  \bibinfo{person}{Sebastian Thrun}.} \bibinfo{year}{2017}\natexlab{}.
\newblock \showarticletitle{Dermatologist-level classification of skin cancer
  with deep neural networks}.
\newblock \bibinfo{journal}{\emph{Nature}} \bibinfo{volume}{542},
  \bibinfo{number}{7639} (\bibinfo{year}{2017}), \bibinfo{pages}{115}.
\newblock


\bibitem[\protect\citeauthoryear{Geirhos, Rubisch, Michaelis, Bethge, Wichmann,
  and Brendel}{Geirhos et~al\mbox{.}}{2019}]%
        {geirhos2018}
\bibfield{author}{\bibinfo{person}{Robert Geirhos}, \bibinfo{person}{Patricia
  Rubisch}, \bibinfo{person}{Claudio Michaelis}, \bibinfo{person}{Matthias
  Bethge}, \bibinfo{person}{Felix~A. Wichmann}, {and} \bibinfo{person}{Wieland
  Brendel}.} \bibinfo{year}{2019}\natexlab{}.
\newblock \showarticletitle{ImageNet-trained {CNN}s are biased towards texture;
  increasing shape bias improves accuracy and robustness.}. In
  \bibinfo{booktitle}{\emph{International Conference on Learning
  Representations}}.
\newblock
\urldef\tempurl%
\url{https://openreview.net/forum?id=Bygh9j09KX}
\showURL{%
\tempurl}


\bibitem[\protect\citeauthoryear{Goodfellow, Shlens, and Szegedy}{Goodfellow
  et~al\mbox{.}}{2014}]%
        {goodfellow2014explaining}
\bibfield{author}{\bibinfo{person}{Ian~J Goodfellow}, \bibinfo{person}{Jonathon
  Shlens}, {and} \bibinfo{person}{Christian Szegedy}.}
  \bibinfo{year}{2014}\natexlab{}.
\newblock \showarticletitle{Explaining and harnessing adversarial examples}.
\newblock \bibinfo{journal}{\emph{arXiv preprint arXiv:1412.6572}}
  (\bibinfo{year}{2014}).
\newblock


\bibitem[\protect\citeauthoryear{Haenssle, Fink, Schneiderbauer, Toberer, Buhl,
  Blum, Kalloo, Hassen, Thomas, Enk, et~al\mbox{.}}{Haenssle
  et~al\mbox{.}}{2018}]%
        {haenssle2018}
\bibfield{author}{\bibinfo{person}{Holger~A Haenssle},
  \bibinfo{person}{Christine Fink}, \bibinfo{person}{R Schneiderbauer},
  \bibinfo{person}{Ferdinand Toberer}, \bibinfo{person}{Timo Buhl},
  \bibinfo{person}{A Blum}, \bibinfo{person}{A Kalloo},
  \bibinfo{person}{A~Ben~Hadj Hassen}, \bibinfo{person}{Luc Thomas},
  \bibinfo{person}{A Enk}, {et~al\mbox{.}}} \bibinfo{year}{2018}\natexlab{}.
\newblock \showarticletitle{Man against machine: diagnostic performance of a
  deep learning convolutional neural network for dermoscopic melanoma
  recognition in comparison to 58 dermatologists}.
\newblock \bibinfo{journal}{\emph{Annals of Oncology}} \bibinfo{volume}{29},
  \bibinfo{number}{8} (\bibinfo{year}{2018}), \bibinfo{pages}{1836--1842}.
\newblock


\bibitem[\protect\citeauthoryear{Howard and Ruder}{Howard and Ruder}{2018}]%
        {howard2018universal}
\bibfield{author}{\bibinfo{person}{Jeremy Howard} {and}
  \bibinfo{person}{Sebastian Ruder}.} \bibinfo{year}{2018}\natexlab{}.
\newblock \showarticletitle{Universal language model fine-tuning for text
  classification}.
\newblock \bibinfo{journal}{\emph{arXiv preprint arXiv:1801.06146}}
  (\bibinfo{year}{2018}).
\newblock


\bibitem[\protect\citeauthoryear{Imaizumi, Watanabe, Hirano, Takemura,
  Kashiwagi, and Monobe}{Imaizumi et~al\mbox{.}}{2017}]%
        {imaizumi2017}
\bibfield{author}{\bibinfo{person}{Hideaki Imaizumi}, \bibinfo{person}{Akio
  Watanabe}, \bibinfo{person}{Hiromi Hirano}, \bibinfo{person}{Masatoshi
  Takemura}, \bibinfo{person}{Hideyuki Kashiwagi}, {and}
  \bibinfo{person}{Shinichiro Monobe}.} \bibinfo{year}{2017}\natexlab{}.
\newblock \showarticletitle{Hippocra: Doctor-to-Doctor TeleDermatology
  consultation service towards future AI-based Diagnosis System in Japan}. In
  \bibinfo{booktitle}{\emph{2017 IEEE International Conference on Consumer
  Electronics-Taiwan (ICCE-TW)}}. IEEE, \bibinfo{publisher}{IEEE},
  \bibinfo{pages}{51--52}.
\newblock


\bibitem[\protect\citeauthoryear{Kimball and Resneck~Jr}{Kimball and
  Resneck~Jr}{2008}]%
        {kimball2008us}
\bibfield{author}{\bibinfo{person}{Alexa~Boer Kimball} {and}
  \bibinfo{person}{Jack~S Resneck~Jr}.} \bibinfo{year}{2008}\natexlab{}.
\newblock \showarticletitle{The US dermatology workforce: a specialty remains
  in shortage}.
\newblock \bibinfo{journal}{\emph{Journal of the American Academy of
  Dermatology}} \bibinfo{volume}{59}, \bibinfo{number}{5}
  (\bibinfo{year}{2008}), \bibinfo{pages}{741--745}.
\newblock


\bibitem[\protect\citeauthoryear{Lanzini, Fallen, Wismer, and Lima}{Lanzini
  et~al\mbox{.}}{2012}]%
        {lanzini2012impact}
\bibfield{author}{\bibinfo{person}{Rosilene~Canzi Lanzini},
  \bibinfo{person}{Robyn~S Fallen}, \bibinfo{person}{Judy Wismer}, {and}
  \bibinfo{person}{Hermenio~C Lima}.} \bibinfo{year}{2012}\natexlab{}.
\newblock \showarticletitle{Impact of the number of dermatologists on
  dermatology biomedical research: a Canadian study}.
\newblock \bibinfo{journal}{\emph{Journal of cutaneous medicine and surgery}}
  \bibinfo{volume}{16}, \bibinfo{number}{3} (\bibinfo{year}{2012}),
  \bibinfo{pages}{174--179}.
\newblock


\bibitem[\protect\citeauthoryear{LeCun, Bengio, and Hinton}{LeCun
  et~al\mbox{.}}{2015}]%
        {lecun2015DL}
\bibfield{author}{\bibinfo{person}{Yann LeCun}, \bibinfo{person}{Yoshua
  Bengio}, {and} \bibinfo{person}{Geoffrey Hinton}.}
  \bibinfo{year}{2015}\natexlab{}.
\newblock \showarticletitle{Deep learning}.
\newblock \bibinfo{journal}{\emph{nature}} \bibinfo{volume}{521},
  \bibinfo{number}{7553} (\bibinfo{year}{2015}), \bibinfo{pages}{436}.
\newblock


\bibitem[\protect\citeauthoryear{Liu, Jain, Eng, Way, Lee, Bui, Kanada,
  Marinho, Gallegos, Gabriele, et~al\mbox{.}}{Liu et~al\mbox{.}}{2019}]%
        {liu2019deep}
\bibfield{author}{\bibinfo{person}{Yuan Liu}, \bibinfo{person}{Ayush Jain},
  \bibinfo{person}{Clara Eng}, \bibinfo{person}{David~H Way},
  \bibinfo{person}{Kang Lee}, \bibinfo{person}{Peggy Bui},
  \bibinfo{person}{Kimberly Kanada}, \bibinfo{person}{Guilherme de~Oliveira
  Marinho}, \bibinfo{person}{Jessica Gallegos}, \bibinfo{person}{Sara
  Gabriele}, {et~al\mbox{.}}} \bibinfo{year}{2019}\natexlab{}.
\newblock \showarticletitle{A deep learning system for differential diagnosis
  of skin diseases}.
\newblock \bibinfo{journal}{\emph{arXiv preprint arXiv:1909.05382}}
  (\bibinfo{year}{2019}).
\newblock


\bibitem[\protect\citeauthoryear{Loshchilov and Hutter}{Loshchilov and
  Hutter}{2016}]%
        {loshchilov2016}
\bibfield{author}{\bibinfo{person}{Ilya Loshchilov} {and}
  \bibinfo{person}{Frank Hutter}.} \bibinfo{year}{2016}\natexlab{}.
\newblock \showarticletitle{SGDR: Stochastic gradient descent with warm
  restarts}.
\newblock \bibinfo{journal}{\emph{arXiv preprint arXiv:1608.03983}}
  (\bibinfo{year}{2016}).
\newblock


\bibitem[\protect\citeauthoryear{Lowell, Froelich, Federman, and
  Kirsner}{Lowell et~al\mbox{.}}{2001}]%
        {lowell2001}
\bibfield{author}{\bibinfo{person}{Brooke~A Lowell},
  \bibinfo{person}{Catherine~W Froelich}, \bibinfo{person}{Daniel~G Federman},
  {and} \bibinfo{person}{Robert~S Kirsner}.} \bibinfo{year}{2001}\natexlab{}.
\newblock \showarticletitle{Dermatology in primary care: prevalence and patient
  disposition}.
\newblock \bibinfo{journal}{\emph{Journal of the American Academy of
  Dermatology}} \bibinfo{volume}{45}, \bibinfo{number}{2}
  (\bibinfo{year}{2001}), \bibinfo{pages}{250--255}.
\newblock


\bibitem[\protect\citeauthoryear{Mishra, Imaizumi, and Yamasaki}{Mishra
  et~al\mbox{.}}{2019a}]%
        {mishra2019interpreting}
\bibfield{author}{\bibinfo{person}{Sourav Mishra}, \bibinfo{person}{Hideaki
  Imaizumi}, {and} \bibinfo{person}{Toshihiko Yamasaki}.}
  \bibinfo{year}{2019}\natexlab{a}.
\newblock \showarticletitle{Interpreting Fine-Grained Dermatological
  Classification by Deep Learning}. In \bibinfo{booktitle}{\emph{Proceedings of
  the IEEE Conference on Computer Vision and Pattern Recognition Workshops}}.
  \bibinfo{publisher}{Computer Vision Foundation}, \bibinfo{pages}{0--0}.
\newblock


\bibitem[\protect\citeauthoryear{Mishra, Yamasaki, and Imaizumi}{Mishra
  et~al\mbox{.}}{2018}]%
        {mishra2018a}
\bibfield{author}{\bibinfo{person}{Sourav Mishra}, \bibinfo{person}{Toshihiko
  Yamasaki}, {and} \bibinfo{person}{Hideaki Imaizumi}.}
  \bibinfo{year}{2018}\natexlab{}.
\newblock \showarticletitle{Supervised classification of Dermatological
  diseases by Deep learning}.
\newblock \bibinfo{journal}{\emph{arXiv preprint arXiv:1802.03752}}
  (\bibinfo{year}{2018}), \bibinfo{pages}{1--6}.
\newblock


\bibitem[\protect\citeauthoryear{Mishra, Yamasaki, and Imaizumi}{Mishra
  et~al\mbox{.}}{2019b}]%
        {mishra2019improving}
\bibfield{author}{\bibinfo{person}{Sourav Mishra}, \bibinfo{person}{Toshihiko
  Yamasaki}, {and} \bibinfo{person}{Hideaki Imaizumi}.}
  \bibinfo{year}{2019}\natexlab{b}.
\newblock \showarticletitle{Improving image classifiers for small datasets by
  learning rate adaptations}.
\newblock \bibinfo{journal}{\emph{arXiv preprint arXiv:1903.10726}}
  (\bibinfo{year}{2019}).
\newblock


\bibitem[\protect\citeauthoryear{Park, Ko, and Swerlick}{Park
  et~al\mbox{.}}{2018}]%
        {park2018}
\bibfield{author}{\bibinfo{person}{Andrew~J Park}, \bibinfo{person}{Justin~M
  Ko}, {and} \bibinfo{person}{Robert~A Swerlick}.}
  \bibinfo{year}{2018}\natexlab{}.
\newblock \showarticletitle{Crowdsourcing dermatology: DataDerm, big data
  analytics, and machine learning technology}.
\newblock \bibinfo{journal}{\emph{Journal of the American Academy of
  Dermatology}} \bibinfo{volume}{78}, \bibinfo{number}{3}
  (\bibinfo{year}{2018}), \bibinfo{pages}{643--644}.
\newblock


\bibitem[\protect\citeauthoryear{Paszke, Gross, Massa, Lerer, Bradbury, Chanan,
  Killeen, Lin, Gimelshein, Antiga, et~al\mbox{.}}{Paszke
  et~al\mbox{.}}{2019}]%
        {pytorch2019}
\bibfield{author}{\bibinfo{person}{Adam Paszke}, \bibinfo{person}{Sam Gross},
  \bibinfo{person}{Francisco Massa}, \bibinfo{person}{Adam Lerer},
  \bibinfo{person}{James Bradbury}, \bibinfo{person}{Gregory Chanan},
  \bibinfo{person}{Trevor Killeen}, \bibinfo{person}{Zeming Lin},
  \bibinfo{person}{Natalia Gimelshein}, \bibinfo{person}{Luca Antiga},
  {et~al\mbox{.}}} \bibinfo{year}{2019}\natexlab{}.
\newblock \showarticletitle{PyTorch: An imperative style, high-performance deep
  learning library}. In \bibinfo{booktitle}{\emph{Advances in Neural
  Information Processing Systems}}. \bibinfo{pages}{8024--8035}.
\newblock


\bibitem[\protect\citeauthoryear{Powles and Hodson}{Powles and Hodson}{2017}]%
        {powles2017google}
\bibfield{author}{\bibinfo{person}{Julia Powles} {and} \bibinfo{person}{Hal
  Hodson}.} \bibinfo{year}{2017}\natexlab{}.
\newblock \showarticletitle{Google DeepMind and healthcare in an age of
  algorithms}.
\newblock \bibinfo{journal}{\emph{Health and technology}} \bibinfo{volume}{7},
  \bibinfo{number}{4} (\bibinfo{year}{2017}), \bibinfo{pages}{351--367}.
\newblock


\bibitem[\protect\citeauthoryear{Selvaraju, Cogswell, Das, Vedantam, Parikh,
  and Batra}{Selvaraju et~al\mbox{.}}{2017}]%
        {selvaraju2017}
\bibfield{author}{\bibinfo{person}{Ramprasaath~R Selvaraju},
  \bibinfo{person}{Michael Cogswell}, \bibinfo{person}{Abhishek Das},
  \bibinfo{person}{Ramakrishna Vedantam}, \bibinfo{person}{Devi Parikh}, {and}
  \bibinfo{person}{Dhruv Batra}.} \bibinfo{year}{2017}\natexlab{}.
\newblock \showarticletitle{Grad-cam: Visual explanations from deep networks
  via gradient-based localization}. In \bibinfo{booktitle}{\emph{Proceedings of
  the IEEE International Conference on Computer Vision}}.
  \bibinfo{pages}{618--626}.
\newblock


\bibitem[\protect\citeauthoryear{Shrivastava, Londhe, Sonawane, and
  Suri}{Shrivastava et~al\mbox{.}}{2015}]%
        {shrivastava2015}
\bibfield{author}{\bibinfo{person}{Vimal~K Shrivastava},
  \bibinfo{person}{Narendra~D Londhe}, \bibinfo{person}{Rajendra~S Sonawane},
  {and} \bibinfo{person}{Jasjit~S Suri}.} \bibinfo{year}{2015}\natexlab{}.
\newblock \showarticletitle{Reliable and accurate psoriasis disease
  classification in dermatology images using comprehensive feature space in
  machine learning paradigm}.
\newblock \bibinfo{journal}{\emph{Expert Systems with Applications}}
  \bibinfo{volume}{42}, \bibinfo{number}{15-16} (\bibinfo{year}{2015}),
  \bibinfo{pages}{6184--6195}.
\newblock


\bibitem[\protect\citeauthoryear{Smith}{Smith}{2017}]%
        {smith2017}
\bibfield{author}{\bibinfo{person}{Leslie~N Smith}.}
  \bibinfo{year}{2017}\natexlab{}.
\newblock \showarticletitle{Cyclical learning rates for training neural
  networks}. In \bibinfo{booktitle}{\emph{2017 IEEE Winter Conference on
  Applications of Computer Vision (WACV)}}. IEEE, \bibinfo{pages}{464--472}.
\newblock


\bibitem[\protect\citeauthoryear{Stern}{Stern}{2010}]%
        {stern2010}
\bibfield{author}{\bibinfo{person}{Robert~S Stern}.}
  \bibinfo{year}{2010}\natexlab{}.
\newblock \showarticletitle{Prevalence of a history of skin cancer in 2007:
  results of an incidence-based model}.
\newblock \bibinfo{journal}{\emph{Archives of dermatology}}
  \bibinfo{volume}{146}, \bibinfo{number}{3} (\bibinfo{year}{2010}),
  \bibinfo{pages}{279--282}.
\newblock


\bibitem[\protect\citeauthoryear{Sun, Yang, Sun, and Wang}{Sun
  et~al\mbox{.}}{2016}]%
        {sun2016}
\bibfield{author}{\bibinfo{person}{Xiaoxiao Sun}, \bibinfo{person}{Jufeng
  Yang}, \bibinfo{person}{Ming Sun}, {and} \bibinfo{person}{Kai Wang}.}
  \bibinfo{year}{2016}\natexlab{}.
\newblock \showarticletitle{A benchmark for automatic visual classification of
  clinical skin disease images}. In \bibinfo{booktitle}{\emph{European
  Conference on Computer Vision}}. Springer, \bibinfo{pages}{206--222}.
\newblock


\bibitem[\protect\citeauthoryear{Suneja, Smith, Chen, Zipperstein,
  Fleischer~Jr, and Feldman}{Suneja et~al\mbox{.}}{2001}]%
        {suneja2001}
\bibfield{author}{\bibinfo{person}{Tina Suneja}, \bibinfo{person}{Edward~D
  Smith}, \bibinfo{person}{G~John Chen}, \bibinfo{person}{Kory~J Zipperstein},
  \bibinfo{person}{Alan~B Fleischer~Jr}, {and} \bibinfo{person}{Steven~R
  Feldman}.} \bibinfo{year}{2001}\natexlab{}.
\newblock \showarticletitle{Waiting times to see a dermatologist are perceived
  as too long by dermatologists: implications for the dermatology workforce}.
\newblock \bibinfo{journal}{\emph{Archives of dermatology}}
  \bibinfo{volume}{137}, \bibinfo{number}{10} (\bibinfo{year}{2001}),
  \bibinfo{pages}{1303--1307}.
\newblock


\end{thebibliography}
\end{document}